\documentclass[a4paper,fleqn]{cas-dc}

\usepackage[utf8]{inputenc}
\usepackage{hyperref}
\usepackage{lineno}
\usepackage{amsmath}
\usepackage{float}
\usepackage{amssymb}
\usepackage{amsfonts}
\usepackage{graphicx}
\usepackage{subcaption}
\usepackage{placeins}
\usepackage{threeparttable}
\usepackage[lined,ruled,vlined]{algorithm2e}
\usepackage{float}
\usepackage{natbib}
\usepackage{etoolbox}

\shortauthors{Naharro et~al.}

\newtheorem{definition}{Definition}

\begin{document}


\title{Comparative study of regression vs pairwise models for surrogate-based heuristic optimisation}
\shorttitle{Comparative study of regression vs pairwise models for surrogate-based heuristic optimisation}

\author[1]{Pablo S. Naharro\corref{cor1}}[orcid=0000-0002-1213-2559]
\ead{p.sanchez@lurtis.com}
\author[2]{Pablo Toharia}[orcid=0000-0003-2429-1300]
\ead{pablo.toharia@upm.es}
\author[2]{Antonio LaTorre}[orcid=0000-0002-8718-5735]
\ead{a.latorre@upm.es}
\author[2,3]{Jos\'{e}-Mar\'{i}a Pe\~{n}a\corref{cor1}}[orcid=0000-0002-3303-0455]
\ead{jm.penya@lurtis.com}

\cortext[cor1]{Corresponding authors}
\address[1]{Lurtis Rules, Calle de los Ciruelos, 28223, Pozuelo de Alarc{\'o}n, Madrid, Spain}
\address[2]{Center for Computational Simulation (CCS), Universidad Polit{\'e}cnica de Madrid, Spain}
\address[3]{Lurtis Ltd., Wood Centre for Innovation Stansfeld Park, Quarry Rd, Oxford, OX3 8SB, UK}

\fntext[fn1]{PT, PSN, and JMP would like to thank the Industrial PhD Programme of \textit{Comunidad de Madrid} (grant IND2019/TIC-17140). PT, ALT, and JMP thank the Spanish Ministry of Science (grants TIN2017-83132-C2-2-R and PID2020-113013RB-C22).}

\begin{abstract}
Heuristic optimisation is a popular tool for solving problems in the sciences and engineering fields. These algorithms explore the search space by sampling solutions, evaluating their fitness, and biasing the search in the direction of promising solutions. However, in many cases, this fitness function involves executing expensive computational calculations, drastically reducing the reasonable number of evaluations. In this context, surrogate models have emerged as an excellent alternative to alleviate these computational problems. This paper addresses the formulation of surrogate problems as both regression models that approximate fitness (surface surrogate models) and a novel way to connect classification models (pairwise surrogate models). The pairwise approach can be directly exploited by some algorithms, such as Differential Evolution, in which the fitness value is not actually needed to drive the search, and it is sufficient to know whether a solution is better than another one or not. 

Based on these modelling approaches, we have conducted a multidimensional analysis of surrogate models under different configurations: different machine learning algorithms (regularised regression, neural networks, decision trees, boosting methods, and random forests), different surrogate strategies (encouraging diversity or relaxing prediction thresholds), and compare them for both surface and pairwise surrogate models. The experimental part of the article includes the benchmark problems already proposed for the SOCO'2011 competition in continuous optimisation and a simulation problem included in the recent GECCO'2021 Industrial Challenge. 

This paper shows that the performance of the overall search, when using online machine learning-based surrogate models, depends not only on the accuracy of the predictive model but also on both the kind of bias towards positive or negative cases and how the optimisation uses those predictions to decide whether to execute the actual fitness function.

\end{abstract}


\maketitle

\section{Introduction}
Heuristic optimisation methods are based on the stochastic exploration of the search space of potential solutions to a problem based on a quality function to be maximised or minimised. In essence, these methods perform an iterative process by testing tentative solutions driven by this quality function and guided by a search method (for instance, population-based evolutionary methods, trajectory methods, Quasi-Newton methods, among many others). Although there is repetitive testing of candidate solutions in the search space, the design of these configurations aims to minimise the number of calls to the fitness function needed to find a good-quality solution to the problem. This is far more efficient than any brute-force or systematic testing alternative. However, there is still a trade-off between the number of candidate solutions explored in the course of the optimisation process and the quality, or even optimality of the search. In theory, \cite{wolpert1997no} suggested that there is no best overall algorithm for every potential problem. However, the research of better and more efficient heuristic optimisation methods has demonstrated that methods perform better or worse depending on the characteristics of real-world problems and non-tailor-made functions (designed to deceive particular search strategies), as stated in \citep{ho2002simple,koehler2007conditions}. Nonetheless, even the best-suited algorithm still requires the evaluation of potential solutions, calculating the result of the quality function and using this value to propose the subsequent solutions to evaluate.

The use of heuristic optimisation in fields such as engineering, science, or finance has shown to be effective in overcoming complex problems \citep{deb2012optimization,tapia2007applications,ponsich2012survey}. However, in many real-world application cases, the evaluation of candidate solutions requires the execution of a simulation or an equivalent expensive computational calculation. For instance, in computational mechanics, the optimisation of material properties or design shapes requires finite element calculations \citep{muelas2008optimizing,dubourg2011adaptive,pena2019softfem}. Other examples are wing design problems, in which the fitness function performs fluid dynamics calculations \citep{moore2016comparison}; or computational neuroscience, in which multiphysics simulations are conducted to calibrate models with experimental results \citep{latorre2020model}. Therefore, making quality function evaluations as computationally efficient as possible is essential in this kind of problems. Hence, some advanced computing techniques to reduce computational time have been proposed, such as CPU multiprocessing or parallel execution \citep{alba2002parallel}, GPU computation \citep{luong2010gpu}, and problem partitioning or multi-scaling injection \citep{sinha2007multiscale}. These approaches require, sometimes, adapting the structure of the heuristic optimisation algorithm in a non-trivial way.

On the other hand, in \citep{jin2002framework} it was proposed to use less intrusive and algorithm-agnostic methods based on the approximation of the fitness function. Among them, surrogate-based optimisation \citep{forrester2009recent,han2012surrogate} is one of the most common approaches in both engineering and science domains.

Surrogate models substitute the quality function by an approximation that can be computed in a much more efficient way, observing a reasonable degree of fidelity compared to the original function. Two approaches for the use of surrogates were analysed in \citep{han2012surrogate}. The surrogate model is either: (1)~calculated initially with the optimisation algorithm using only the surrogate as the quality function, or (2)~surrogates are updated during the optimisation that alternatively uses the surrogate or the actual quality function according to a given strategy.

In the first approach, once these points are selected and the corresponding values of the quality function calculated, they are approximated by a model (surrogate). This model is ultimately the objective function optimised by either classical gradient-based methods or heuristic optimisation (e.g., a genetic algorithm). In the second option, when the optimisation also updates the surrogate model, it is necessary to define a strategy that alternates between using the surrogate to predict the output of the fitness function and using the actual fitness evaluation. Within this group, the concept of filtering is proposed by \cite{Berveglieri2020}, using the surrogate model to make a prediction of the fitness function; if this solution is promising, then the actual fitness function is calculated. This approach always uses the original quality function, but the surrogate approximation allows the optimisation to discard non-promising solutions. As the optimisation process collects new actual points, it is possible to update the surrogate model and refine the approximation. 

Moreover, the traditional approach uses a regression problem as a surrogate that replaces the real problem unless a certain criterion is met. Several proposals have been made using Polynomial Models, Kriging Models, Radial Basis functions (RBF), Artificial Neuronal Networks (ANNs), Support Vector Regressors (SVRs), Gaussian Processes, as well as a ensemble models \citep{Jin2005,Talbi2021}.

However, for multiple algorithms (local search algorithms and specific population-based optimisation techniques), it is less important to know the actual fitness value of new candidate solutions than whether they are better or worse than another reference solution. Thus, this approach makes it possible to reformulate the problem in terms of a classification problem from the perspective of the machine learning algorithm that implements the surrogate. Different approaches have been proposed within classification techniques. Some of them aim to reduce complexity by discretising the problem. A level-based classifier that divides the trials into 4 discrete layers was proposed by \cite{Wei2021} in combination with a Particle Swarm Optimisation (PSO). The algorithm by \cite{Shen2016} generates offspring candidates with multiple operators and uses a classifier to select which operator is going to be used for that specific trial. Other proposals are focused on the elitism criteria used by the optimisation algorithm to connect the classifier. CADE \citep{Lu2011} proposes the use of a classifier that predicts whether the sample will enter the new population or not. For this, a model is trained every generation using pairwise neighbourhood data. Furthermore, CRADE \citep{Lu2012}, an extension of CADE, combines regression and classification methods, outperforming DEGL \citep{Das2009}, JADE \citep{Zhang2009} and CoDE \citep{Wang2011}. It also suggests cutting the execution under 10.000 fitness evaluations since it can be prohibitive for too time-consuming real-world problems. Moreover, authors such as \cite{Liu2021} have recently performed a comparison between regression and classification models, but only in an offline data-driven EA. 

This paper includes a comparison of the online pairwise classification alternative versus the online classical regression approach to implement the surrogate, analysing different machine learning algorithms (regularised regression, neural networks, decision trees, boosting methods, and random forests). Additionally, the selection of supervised machine learning models (used in either regression or their counterparts in classification) is complemented with experiments on different improvements in the updating strategy and the conditions under which the actual quality function is applied. 

According to \cite{Berveglieri2020}, the choice of a particular surrogate can be critical, and our objective is to analyse several machine learning algorithms and strategies to design surrogates. In order to achieve this, the optimisation algorithm is fixed to a standard yet competitive performing method, a standard differential evolution (DE) \citep{price2006differential}. This way, it is possible to keep the complexity of the optimisation algorithm fixed, focusing the scope of the paper on the surrogate design and alternatives. DE is a recurrent base algorithm in most of the hybrid and self-adaptive methods that have dominated the most recent continuous optimisation competitions in mid- to large-scale problems \citep{latorre2011mos,latorre2015comprehensive,latorre2017comparison}. DE performance is by itself quite competitive compared to Genetic Algorithms (GA) or Estimation of Distribution Algorithms (EDA). 
 
Also, according to authors such as \cite{lim2009generalizing} ``\textit{observations obtained from the use of benchmark problems do not deviate significantly from those in the real-world problems we have experimented with}''. Therefore, the results of benchmark problems can be of value for real-world problems. Other authors, such as \cite{Rehbach2018}, also support this conclusion. For this reason, the paper includes an extended experimental section using the functions proposed by \cite{lozano2011editorial} for the SOCO'2011 special issue, a well-known continuous global optimisation benchmark.

The remainder of this paper is organised as follows. Section~\ref{sec:related} presents relevant related work to the topics discussed in this paper. In Section~\ref{sec:contrib}, we present the design of the study, including the machine learning algorithms and all the optimisation strategies under comparison. In Section~\ref{sec:pairwise}, a novel approach that formulates the surrogate modelling as a binary classification instead of a regression problem is presented. The experimental scenario (benchmark, statistical validation, and hyper-parameters tuning) is depicted in Section~\ref{sec:experiments}, whereas Section~\ref{sec:results} presents the results obtained. A thorough discussion of these results is presented in Section~\ref{sec:discussion}.
In order to contrast the performance against a real-world problem, Section~\ref{sec:realworld} includes the results of a simulation problem proposed in the recent GECCO 2021 Industrial Challenge. Finally, Section~\ref{sec:conclusiones} concludes this work by highlighting the most relevant insights obtained during this study. 

\section{Related Work}\label{sec:related} 
Surrogate-based optimisation \citep{jin2011surrogate} is an active field, in which an optimisation method (typically a heuristic method, such as evolutionary algorithms or local search methods) uses a model (named surrogate) to approximate the fitness function. This surrogate can be used to optimise the problem instead of the actual fitness function (making the search method to explore over this approximate fitness landscape) or can be used online during the optimisation to alternatively use either the surrogate or the actual fitness function \citep{nikolos2013use}.

Surrogate-based optimisation has been extensively used when the fitness function is a simulation, such as finite element models \citep{dubourg2011adaptive,awad2018improved,pena2019softfem}, fluid dynamics \citep{zhou2006combining,forrester2006optimization,nikolos2013use,moore2016comparison}, or energy efficiency models \citep{nguyen2014review,westermann2019surrogate}.

The most common surrogate models use Design of Experiments (DoE) methods to preliminary sample the solution space using a limited number of points (typically sampling by using strategies such as Orthogonal Array Design, Latin Hypercube Sampling, or Sobol Sequences; \citep{giunta2003overview}). Once these points are sampled, surrogates are modeled by using Gaussian Processes (GP), also known as Kriging methods \citep{ong2003evolutionary,zhou2006combining,forrester2006optimization,lim2009generalizing,dubourg2011adaptive,diaz2016review,awad2018improved}, or Response Surface Methodology (RSM) \citep{diaz2016review,yang2016application}. However, and in particular for those methods updating the surrogates during the execution of the optimisation algorithm, different machine learning methods have also been proposed; methods such as artificial neural networks (ANN) \citep{nikolos2013use,moore2016comparison}, Radial Basis Kernels (RBK) \citep{ong2003evolutionary,lim2009generalizing,diaz2016review}, K-nearest neighbours (KNN) \citep{moore2016comparison}, n-order polynomial regression \citep{lim2009generalizing,diaz2016review}, or Support Vector Machines/Regressors (SVM/R) \citep{mallipeddi2015evolving}. Moreover, due to the wide variety of possibilities when connecting machine learning models with the optimisation algorithm, \cite{Talbi2021} proposes a taxonomy to define how the model drives the process.

Despite the fact that some surrogate-based optimisations use trajectory or even non-stochastic techniques \citep{forrester2006optimization}, most of the typical approaches are based on evolutionary methods and tend to use genetic algorithms. However, more recent approaches have also included other metaheuristics, such as Differential Evolution (DE) \citep{nikolos2013use,mallipeddi2015evolving,awad2018improved}, Particle Swarm Optimisation (PSO) \citep{nguyen2014review}, and hybrid/memetic approaches \citep{ong2003evolutionary,zhou2006combining,lim2009generalizing,nguyen2014review}. There is also some relevant literature in multi-objective optimisation \citep{lim2009generalizing,yang2016application}.

Different metrics to compare the performance of surrogate algorithms were proposed by \cite{moore2016comparison}, and they address four different criteria: (1)~accuracy, (2)~efficiency, (3)~ robustness, and (4)~performance.
\begin{enumerate}
\item Accuracy is measured as the error between predictions from the model compared to the actual result of the fitness function. These are typically measures such as the mean square error (MSE) or the coefficient of determination ($R^2$).
\item Efficiency represents how many samples are needed for the surrogate to approximate the fitness function.
\item Robustness measures the variability in the performance of the algorithm when varying the initial conditions. It is typically measured by analysing several runs of the algorithm with different random seeds.
\item Performance, which is the final goal, measures both the time and the value of the best fitness found by the algorithm after the stop conditions. In this field, \cite{Rehbach2018} criticises the way the performance is measured when enabling parallelism. Meanwhile, pure sequential executions are likely to get good results with a low number of fitness function evaluations; parallel executions are prone to reduce the execution time but need more evaluations. 
\end{enumerate}

Finally, in \citep{viana2010making}, the design principle to construct effective surrogate models is analysed , which includes (1)~feature selection (either explicitly feature filtering or by using machine learning algorithms that perform internal feature selection or regularisation, sometimes even a sensibility analysis \citep{nguyen2014review}), (2)~consideration of different surrogate models (simultaneously, dynamically electing them or by ensemble or competition), and (3)~the use of conservative estimators. This last point poses the question whether the criteria for selecting a surrogate model differ from the accuracy measures typically used in machine learning. In surrogate-based optimisation, the kind of errors derived from the approximate estimation of the fitness function do not have the same impact on the overall performance of the algorithm. This is very clear when the surrogate is biased towards overestimation or underestimation (assuming a minimisation problem), and results in a lack of search guidance or no significant computation reduction, respectively. To avoid these last problems, there are proposals that reduce performance with the goal of getting a better result. In addition, a method to use both model prediction and its confidence bound was proposed by \cite{hagg2020airflow}.

\section{Design of the Study}
\label{sec:contrib}

The objective of this analysis is to compare the effectiveness of different machine learning algorithms to model a surrogate that will be updated during the optimisation. The selection of these algorithms is detailed in Section~\ref{sec:ml}. A second aspect under analysis is how the optimisation method uses the surrogate to decide whether it is interesting or not to calculate the actual values of the quality function for a given candidate solution. This decision process not only has an impact on the effectiveness of the surrogate model (to save quality function evaluations by discarding solutions that would not improve the current best solution), but also on the collection of additional data points to update the surrogate model itself. Section~\ref{sec:strategies} proposes a baseline strategy for the application of surrogates and Section~\ref{sec:combined} three complementary criteria that can be combined to create up to eight different strategies. The complete analysis uses all these strategies in conjunction with each of the machine learning models to identify synergies that perform better depending on the characteristics of the machine learning technique to tackle the quality function modelling.

Following the taxonomy proposed in \citep{Talbi2021}, this paper uses a population-based evolutionary algorithm in combination with a low-level data-driven approach. This is achieved by introducing a machine learning layer as a surrogate model in the solution recombination process.

\subsection{Machine Learning Methods}\label{sec:ml}
Several machine learning algorithms have been used as model surrogates. This study has selected five of them that represent a broad range in both complexity and expressive power to capture the quality function characteristics. Although there are more advanced versions of these algorithms in the literature, for the sake of generality, this study includes only the most common ones. The rationale behind this selection is to focus on the aspects of these models that complement the different strategies for the decision process that uses them as surrogates.

The algorithms used for surrogate modelling are the following:

\begin{enumerate}
    \item Ridge Regression (Ridge/R): A regression model that minimises the linear least square error using L2-norm regularisation \citep{hoerl1970ridge}. 
    \item Multilayered Perceptron Regressor(MLP/R): A 100-neuron hidden layer network activated by a rectified linear unit function and solved by square-loss minimisation using a stochastic gradient-based descender \citep{kingma2014adam}.
    \item Decision Tree Regressor (DT/R): A decision tree model using feature selection criteria for the splitting based on variance reduction and L2 loss minimisation as defined in \citep{breiman1984classification}.
    \item Random Forest Regressor (RF/R): A meta-regressor proposed in \citep{breiman2001random} that fits 40 sub-samples of the dataset into regression trees.
    \item XGBoost Regressor (XGBoost/R): A parallel tree gradient booster ensemble based on the framework proposed in \citep{friedman2001greedy} in the implementation provided in \citep{chen2016xgboost} as a regressor.
\end{enumerate}

\subsection{Surrogate-based Strategies}\label{sec:strategies}
The objective of using a surrogate is to predict, in an approximate way, the result of the quality function without executing it. Although there are surrogate-based optimisers that directly optimise this surrogate model, this study addresses the approach that applies the surrogate in conjunction with the actual quality function. This approach, as it keeps executing the quality function of new candidate solutions, generates additional data points. Therefore, the optimisation process can use these new data points to update the surrogate model on-the-fly and refine its results.  

This study has established that the surrogate model update occurs at the end of every generation of the evolutionary algorithm. Section~\ref{sec:experiments} describes the specific implementation details and parameters that have been used for the experimental setup. 

\subsubsection{Baseline Algorithm and Default Strategy}\label{sec:baseline}
DE also has a characteristic that makes it interesting for this study. Evolutionary or population-based optimisation algorithms maintain a set of candidate solutions (population) that evolves during the optimisation process. Generally, these algorithms rank these solutions according to the value of the quality function to perform operations such as selection or replacement. However, in DE, each of the candidate solutions is compared with a newly produced solution (the challenger) to test if it improves its current value. If so, the new solution replaces the old one, and this process is repeated at every generation. This means that DE does not need to rank the solutions on the population. Instead, it compares current solutions with the corresponding challenger. This one-to-one comparison allows surrogate-based optimisation to have a criterion to discard candidate solutions if the surrogate estimates that the challenger is not improving the quality of the current solution.

Let $q: \mathbb{X} \to \mathbb{R}$ be the quality function to optimise. The objective of the optimisation is to find $x^*$ such as:
\begin{equation}
\nexists x \in \mathbb{X} : q(x) \prec q(x^*) \label{eq:auality}
\end{equation}

Let $\mathbb{X}$ be the solution space and $\prec$ the binary relation ``\textit{being better than}'' that compares quality function values (if the goal of the optimisation problem is to minimise the quality function $\prec\ \equiv \ <$). 
The \textit{baseline algorithm} (Algorithm~\ref{alg:baseline}) takes a regression-based surrogate $\hat{q}: \mathbb{X} \to \mathbb{R}$ to model the surface of the quality function. This surrogate $\hat{q}$ shall predict the real-valued result of the quality function by fitting the data points of past solutions and their corresponding quality values; and a good surrogate should approach: 
\begin{equation}
\forall x \in \mathbb{X} : \hat{q}(x) \approx q(x)
\end{equation}

\begin{algorithm}[!htp]
    \SetKwFunction{FCalculate}{IsAcceptedByFilter}%
    \SetKwFunction{FReplaceInPopulation}{ReplaceInPopulation}%
    \SetKwFunction{FInsertInTrainingSet}{InsertInTrainingSet}%
	\SetAlgoLined 
        \KwData{A current solution $x$ and a challenger solution $x'$} 
        \KwResult{The best one from both $x$ and $x'$ is kept in the population} 
        \eIf{\FCalculate{$x$, $x'$}}{
             \tcp{The challenger seems to }
             \tcp{improve the current solution}
             \tcp{Therefore, the actual quality}
             \tcp{ function is computed}
             $v \leftarrow q(x')$\;
             \eIf {$v \prec q(x)$}{
                   \tcp{The actual quality value of}
                   \tcp{$x'$ is better than $x$}
                   \FReplaceInPopulation{$x$, $x'$}\;}{
                   \tcp{The actual quality value is}
                   \tcp{not better}}
            \FInsertInTrainingSet{$x'$, $v$}\;
        }{
        \tcp{Challenger solution $x'$ is}
        \tcp{discarded}
        }
     \caption{Baseline algorithm\label{alg:baseline}.} 
\end{algorithm} 

The \textit{default strategy} (Algorithm~\ref{alg:default}) decides whether to calculate the challenger or discard it, depending on the prediction of the surrogate about the quality function and if it improves the current solution or not, \textit{i. e., Being A the current (challenged) solution and B the new (challenger) solution. If the surrogate estimates that B is better than A, then B is real calculated.}

\begin{algorithm}[!htp]
        \KwData{A current solution $x$ and a challenger solution $x'$} 
        \KwResult{Whether the challenger solution $x'$ is calculated} 
        \SetKwFunction{FCalculate}{IsAcceptedByFilter}%
        \SetKwProg{Fn}{def}{ is}{end}
        \Fn{\FCalculate{$x$, $x'$}}{
           \tcp{baseline calculate/discard }
           \tcp{function}
           \Return $\hat{q}(x') \prec q(x)$
        }
     \caption{Default calculate/discard strategy\label{alg:default}.} 
\end{algorithm}   

In the following sections, we review some alternative strategies for the calculation of the actual quality function with different objectives. These strategies are then compared during the experimentation phase as shown in Section \ref{sec:experiments}.

\subsubsection{Probability-based Strategy}
With the default strategy mentioned above, the optimisation algorithm only calculates the actual quality function if the surrogate prediction indicates that it would improve the existing solution in the population. However, this approach may lead to poor performance if the surrogate overestimates the quality of the solutions. In this case, the surrogate-based strategy will not affect the optimisation process as it would evaluate all candidate solutions (there would be no solution filtering). The opposite effect is also possible, as this approach may cause stagnation if the surrogate were biased towards low-quality estimations. This bias causes the optimisation algorithm to discard all new candidates. The latter of these situations is more likely when the first surrogate has not had enough information to model those areas close to the optimal value, where high-quality solutions are located. 

Algorithm ~\ref{alg:prob} shows a simple variation of the default selection/discard strategy used by the baseline algorithm which incorporates an additional condition that selects a challenger for evaluation. This new condition will be modelled based on a Bernoulli distribution $X \sim Bernoulli(p)$ with coefficient $p=0.2$ regardless of the estimation cast by the surrogate, \textit{i. e., Being A the current (challenged) solution and B the new (challenger) solution. If the surrogate estimates that B is better than A, then B is real calculated. Otherwise, B still have a 0.2 probability of being real calculated.}
 
 \begin{algorithm}[!htp]
        \KwData{A current solution $x$ and a challenger solution $x'$} 
        \KwResult{Whether the challenger solution $x'$ is calculated} 
        \SetKwFunction{FCalculate}{IsAcceptedByFilter}%
        \SetKwProg{Fn}{def}{ is}{end}
        \Fn{\FCalculate{$x$, $x'$}}{
           \tcp{probability-based}
           \tcp{calculate/discard function}
           \Return $\hat{q}(x') \prec q(x) \quad \lor \quad Bernoulli(p)$==true
        }
     \caption{Probability-based calculate/discard strategy\label{alg:prob}.} 
\end{algorithm}

The value $p=0.2$ is itself, a new hyperparameter that is susceptible to be tuned. In our explorative analysis, we have tested several values for this hyperparameter and set it to this value, in which strategies shows differences in their behaviour compared to the default strategy version and without masking surrogate effect.  For the sake of simplicity, this value will remain constant across all strategies as all of them rely on the probability of bypassing the surrogate model decision. 

\subsubsection{Quality Distance Strategy} 
The probability-based strategy provides additional solutions by randomly selecting challengers independently of the estimated quality. Although this strategy ensures a mechanism to avoid stagnation induced by quality underestimation, the surrogate would update the model with potentially low-quality solutions. Alternatively, if the optimisation algorithm wanted to improve the surrogate with new solutions, it would instead need to select those with expected better quality.

This strategy proposes an alternative criterion that will give a second chance to some solutions to be evaluated with the actual quality function. It is based on how sure the surrogate is, that whether the predicted quality of the challenger is not going to improve the current candidate solution. Let $x$ be the current solution, and $x'$ the challenger. Hence, this strategy is controlled by the difference $d(x,x')$ between the estimated value provided by the surrogate and the predicted quality value of the current solution.    
\begin{equation}
    d(x,x') = |\hat{q}(x)-\hat{q}(x')|
    \label{eq:pred}
\end{equation}

The range of values that this difference $d(x,x')$ calculates will vary in the course of the optimisation. In order to bring it to scale, a correction based on the average values of these differences is applied:

\begin{equation}
    \overline{d} = \frac{\sum_{\langle x, x' \rangle}^{P_i} d(x,x')}{card(P_i)}
    \label{eq:pred-average}
\end{equation}

Where $\overline{d}$ is the mean of all differences between the pairs $\langle x,x' \rangle$ of comparisons belonging to $P_i$, where $P_i$ represents the set of all pairwise comparisons the algorithm has checked up to the instant $i$. Let $card(P_i)$ be the cardinality of that set of pairs.
The value of $\overline{d}$ shall be calculated at each iteration of the original Algorithm~\ref{alg:baseline} as more pairs $\langle x, x' \rangle$ are included into the set $P_{i+1} = P_i \cup \{ \langle x, x' \rangle \}$.

This \textit{quality distance strategy} uses the average pairwise distance $\overline{d}$ to establish the probability for a solution to get selected based on the distance in the quality space. This probability is provided by the function $p_{q}(x, y)$,  which shall be of the form: 
\begin{equation}
	p_{q}(x, y) = b^{-d(x,y)} \label{eq:qual-prob}
\end{equation}

To calibrate the base $b$ in the function $p_{q}(x, y)$, a reference point corresponding to the average pairwise distance $\overline{d}$ is set to probability $0.2$:

\begin{equation}
	p_{q}(x, y) = 0.2  \quad \forall x, y \in \mathbb{X} :  d(x,y)=\overline{d}
	\label{eq:qual-calib}
\end{equation}

Therefore, applying Equation~\ref{eq:qual-prob} to the calibration condition set by Equation~\ref{eq:qual-calib}, the value of $b$ is determined as:

\begin{align}
    b^{-\overline{d}} = 0.2                 & {\buildrel \log_{0.2} \over {\quad\Longrightarrow\quad}} \\
    \log_{0.2}(b) = \frac{1}{-\overline{d}} & {\buildrel pow(0.2)   \over {\quad\Longrightarrow\quad}} \\
    b = 0.2^{-\frac{1}{\overline{d}}} &
    \label{eq:qual-base}
\end{align}

The function $p_{q}(x, y)$ has the following properties:
\begin{align}
    \lim_{d(x,y)\to 0}      p_{q}(x,y) & \to 1.0 \\
    \lim_{d(x,y)\to \infty} p_{q}(x,y) & \to 0.0 
\end{align}

This \textit{quality distance strategy} is controlled by the same \texttt{IsAcceptedByFilter} condition shown in Algorithm~\ref{alg:prob}. When, for a new pair of current and challenger solutions $\langle x, x' \rangle$, the probability $p$ of the Bernoulli distribution is set as:

\begin{equation}
    p = \left\{
    \begin{array}{ll}
        1,            & \quad \hat{q}(x') \prec q(x) \\
        p_{q}(x, x')  & \quad \text{otherwise} 
    \end{array}
    \right.
    \label{eq:qual-prob-final}
\end{equation}

\textit{i. e., Being A the current (challenged) solution and B the new (challenger) solution. If the surrogate estimates that B is better than A, then B is real calculated. Otherwise, A and B predictions are passed to a function that determines the probability of B being real calculated. When difference between A and B is smaller, probability of B being calculated tends to 1. The higher the distance is, the lower the probability.}

\subsubsection{Candidate Diversity Strategy}
The previous strategy helps to calculate challenger solutions when the estimation of its quality is close to the threshold set by the quality of the current solution. Hence, the optimisation may at least evaluate and eventually include promising solutions, if the estimation error is reasonably low. However, the \textit{quality distance strategy} does not necessarily improve the quality of the surrogate, as the new solutions that it evaluates are with high probability very close to those already used in training. The \textit{candidate diversity strategy} proposes to select solutions for evaluation (they will be either included or excluded from the population depending on the regular conditions of DE selection) based on the diversity they provide in the space of candidate solutions. The more different the new solution is, compared to the ones already evaluated, the higher the probability of calculating it becomes. 

This strategy is meant to encourage the exploration of other areas of the search space. The optimisation algorithm is the one eventually generating the candidate solutions in those unexplored areas. However, if the estimation of the quality is not good enough to suggest its calculation, this strategy helps in the analysis of the undersampled regions of the search space.   

Let $x$ be a current solution, $S_i \subset \mathbb{X}$ the set of solutions already calculated at the instant $i$, and $x'$ the challenger solution. The function $v: \mathbb{X} \to \mathbb{R}$ is defined as:

\begin{equation}
	v(x', S_i) = min_{x\in S_i}\Vert x-x' \Vert \label{eq:diver}
\end{equation}

Let $\Vert y \Vert$ be a norm function defined on the vector space of the potential solutions $\mathbb{X}$. In particular, the L2 norm (euclidean distance) has been used for this study. The function $v(x', S_i)$ represents the distance from the challenger $x'$ to the closest point in the set of already calculated solutions $S_i$.

As in the case of the previous strategy, the average value of this metric is used to normalise it during the optimisation. 

\begin{equation}
    \overline{v} = \frac{\sum_{(x)}^{S_i} v(x, S_i/\{x\})}{card(S_i)}   \label{eq:diver-average}
\end{equation}

The value of $\overline{v}$ is updated at every iteration and whenever a new solution is added to the set of already calculated solutions $S_{i+1} = S_i \cup \{x\}$. 

This strategy uses the average divergence $\overline{v}$ to establish the probability to get selected based on the distance in the vector space of the potential solutions $\mathbb{X}$. This probability is set by the function $p_{v}(x,S)$, which shall be of the form:

\begin{equation}
    p_{v}(x, S) = 1 - b^{-v(x, S)} \label{eq:diver-prob}
\end{equation}

As in the previous strategy, the base $b$ is calibrated to make $p_{v}(x, S)=0.2$ for those challenger points with distance equal to the average divergence $\overline{v}$. 

\begin{equation}
    p_{v}(x, S) = 0.2  \quad \forall x \in \mathbb{X}, S \subseteq  \mathbb{X}:  v(x, S)=\overline{v}
    \label{eq:diver-calib}
\end{equation}

Applying Equations~\ref{eq:diver-prob} to the calibration condition imposed by Equation~\ref{eq:diver-calib}, the value of $b$ is determined as:

\begin{align}
    1 - b^{-\overline{v}} = 0.2 
    & {\buildrel \log_{0.8} \over {\quad \Longrightarrow \quad}} \\
    \log_{0.8}(b) = \frac{1}{-\overline{v}}
    & {\buildrel pow(0.8) \over {\quad \Longrightarrow \quad}} \\
    b = 0.8^{-\frac{1}{\overline{v}}}
    \label{eq:diver-base}
\end{align}

Opposite to the previous case, this function $p_{v}(x', S_i)$ has the following properties:

\begin{align}
    \lim_{v(x,S)\to 0}      p_{v}(x,S) & \to 0.0 \\
    \lim_{v(x,S)\to \infty} p_{v}(x,S) & \to 1.0 
    \label{eq:diver-props}
\end{align}

Finally, this \textit{candidate diversity strategy} is controlled by the same \texttt{IsAcceptedByFilter} condition shown in Algorithm~\ref{alg:prob}. When, for a new challenger solutions $x'$ and the current set of already calculated solutions $S_i$, the probability $p$ of the Bernoulli distribution is set as:

\begin{equation}
    p = \left\{\begin{array}{ll}
        1,            & \quad \hat{q}(x') \prec q(x) \\
        p_{v}(x', S_i)  & \quad \text{otherwise}
    \end{array}
    \right.
    \label{eq:diver-prob-final}
\end{equation}

\textit{i. e., Being A the current (challenged) solution and B the new (challenger) solution. If the surrogate estimates that B is better than A, then B is real calculated. Otherwise, the distance between B and the nearest neighbour already calculated is passed to a function that determines the probability of B being real calculated. When the distance is small, probability of B being calculated tends to 0. The higher the distance is, the higher the probability.}

\subsection{Combined Strategies}\label{sec:combined}
This article aims to analyse the performance of the different machine learning models, listed in Section~\ref{sec:ml}, as surrogates.  This analysis also includes the application of the different strategies mentioned before, independently and in combination (see Table~\ref{tab:strategy-combos}).

\begin{table}[!htp]
    \centering
    \begin{tabular}{ll}
        \hline
        \textbf{Label} & \textbf{Strategy}\\
        \hline
    	Baseline & Any default strategy \\
        Prob.     & Probability-based Strategy \\
        Qual.     & Quality Distance Strategy \\
        Diver.    & Candidate Diversity Strategy \\
        \hline    
    \end{tabular}
    \caption{Summary of the different strategies.}
    \label{tab:strategy-combos}
\end{table}

All the combinations implement the default strategy and potentially, one or more of the other strategies. In the cases in which multiple strategies are combined, the condition to calculate a challenger solution is the disjunction of all the \texttt{IsAcceptedByFilter} conditions (logical or-clause). If any of the conditions are met, the challenger is calculated.

The experiments (Section~\ref{sec:experiments}) will analyse all the eight combinations of the three strategies to determine their individual contribution and their potential synergies. 

These experiments, using any continuous variable supervised learning (regression) model, are referred hereinafter as \textit{surface surrogates}.

\section{Pairwise Surrogate Model}\label{sec:pairwise}
Most of the surrogate-based literature approaches the creation of models as a continuous variable \textit{surface surrogate} (as a supervised learning problem using regression). There are few cases in which these models are designed as discrete variable supervised learners (classifiers), and mostly only in multiobjective optimisation \citep{pan2018classification} to predict solution dominance using artificial neural networks. However, DE and other similar algorithms compare solutions in a pairwise way. A newly generated candidate solution is compared to a reference solution. These methods do not need to rank the solutions on the population. They only need to determine if a given solution is better or not than a reference solution. This same feature is part of local search and trajectory methods: these methods keep just one working solution that is iteratively updated and compared with neighboring solutions. 

If the optimisation algorithms perform this pairwise comparison as a mechanism to carry out solution selection and replacement, there is an alternative way to model the surrogate as a pairwise classification problem. If this comparison between two solutions is not part of the mechanism of the optimisation algorithm, the application of the pairwise surrogate model is not as straightforward as in the case of DE or local searches. However, there are indirect mechanisms to use this approach in other algorithms. For instance, GA may use tournament selection, a pairwise operation by definition. If the algorithm implements any form of elitism to select whether a solution is included in the population or not, the reference solution is the worst current solution already evaluated. Nonetheless, even though there could be no direct application of the approach and it is needed to implement it indirectly, the advantage of augmenting the ratio of data points for training per fitness evaluation is an interesting benefit for applying this surrogate approach to a problem. This paper is not going to address these issues, which are open for future research.

Let $w: \mathbb{X} \times \mathbb{X} \to \{0, 1\}$ be a boolean function defined in the domain of the pairs of potential solutions $\mathbb{X}$:

\begin{equation}
   w(x,y) = \left\{
    \begin{array}{ll}
        1,  & \quad q(y) \prec q(x) \\
        0,  & \quad \text{otherwise} \\
    \end{array}
    \right.
    \label{eq:pairwise-base}
\end{equation}

The function $w(x,x')$ controls the selection of solutions in DE (and other similar algorithms) when $x$ is the current solution and $x'$ is the corresponding challenger solution. When $w(x,x')$ returns 1, $x'$ value is selected and therefore will replace $x$. Otherwise, $x'$, the challenger, is discarded. Consequently, pairwise surrogate $\hat{w}(x,x')$ is defined as an estimator of the function $w(x,x')$.

\begin{definition}
The boolean function $\hat{w}: \mathbb{X} \times \mathbb{X} \to \{0, 1\}$ is the \textbf{pairwise surrogate}, defined by the selection control $w(x,y)$, as the function that approaches:
\begin{equation}
\forall x,y \in \mathbb{X} : \hat{w}(x,y) \approx w(x,y)
\end{equation}
\end{definition}

Its implementation keeps using Algorithm \ref{alg:baseline} but filter is replaced by Algorithm \ref{alg:pairwise}. \textit{i. e., Being A the challenged and B the challenger. The pairwise surrogate model estimates whether if B should be calculated or not.}

\begin{algorithm}[!htp]
        \KwData{A current solution $x$ and a challenger solution $x'$} 
        \KwResult{Whether the challenger solution $x'$ is calculated} 
        \SetKwFunction{FCalculate}{IsAcceptedByFilter}%
        \SetKwProg{Fn}{def}{ is}{end}
        \Fn{\FCalculate{$x$, $x'$}}{
           \tcp{pairwise surrogate estimation for calculating/discarding }
           \tcp{function}
           \Return $\hat{w}(x,x')$ 
        }
     \caption{Pairwise surrogate estimation for deciding whether to evaluate or discard the solution  \label{alg:pairwise}.} 
\end{algorithm}


This proposed change in the formulation of a surrogate approximation has several interesting considerations:
\begin{enumerate}
\item It is possible to use discrete variable supervised learners (classifiers) as well as continuous variable supervised learners (regressors). In the case of classifiers, they shall model a binary prediction problem. Therefore, there is a broad number of methods to solve these problems (e.g., decision trees, both deep and shallow neural network, Bayesian classifiers, support vector machines,  to name a few). Binary classification is one of the problems that machine learning literature has more extensively treated.  Moreover, regression-based methods can also be applied for binary classification if an acceptance threshold is defined.
\item Binary classification is, in most cases, easier to model than the equivalent regression counterpart. Whereas continuous regression has to fit the model to all the points to minimise an error function, for basic binary classifiers, the task is to identify the border that separates both classes. In addition, some problems are linearly separable (and, therefore, easier for a simple classifier) but it is harder to fit a linear model that predicts those same points. 
\item Surrogate modelling is very constrained by the number of solutions that can be computed at a given time or with a given amount of resources. In order to make surrogates efficient, the number of samples required to build a good model (data points for which the actual quality function has been computed) shall try to be minimised. For a sample of $N$ data points, a binary classifier can train on $N \times N$ data points by creating all the pairwise combinations of those $N$ samples. Pairwise surrogate models may take a greater profit from a reduced number of samples. This aspect might be a problem for the cases collecting a larger number of samples as the quadratic combinations can easily scale to huge numbers making the training process more computationally expensive, but there is always the possibility to undersample those potential combinations up to the trade-off value that makes a compromise between model accuracy and computational training cost. 
\item The training dataset in the case of creating pairwise combinations not only squares the number of data points (instances) but also duplicates the number of features (dimensions), $2 \times D$, with $D$ being the number of components/variables of the solution. Moreover, it is also possible to generate new features that calculate the relationships between the two samples by component, for instance, the difference component-by-component between the two samples (the vector from one solution to the other).  Having more features is not always intrinsically better for classification; it could even be counterproductive if these features are redundant or non-informative. However, there are feature selection mechanisms that may help in preprocessing these training datasets to pick only those relevant features. 
\end{enumerate}

\subsection{Training Pairwise Surrogates}\label{sec:pairwise-mapping}
This paper introduces the use of \textit{pairwise surrogate models} as an alternative to approximate the selection process in a heuristic optimisation problem instead of using an approximation of the quality function. The proposed approach trains a binary classifier on a dataset made of all the pairwise combinations of the available data points to predict whether the first of the components of the pair is better than the second.

Let $(P,q)$ be the training data when $P \subseteq \mathbb{X}$ is the set of solutions available to model the surrogate and $q$ the quality function as defined by Equation~\ref{eq:auality}. 

\begin{definition}
$(W,b)$ is the equivalent \textbf{pairwise training dataset} for $(P,q)$ defined as $W$ the set of pairwise combinations and $b$ the boolean selection function defined on $b: W \to \{0, 1\}$:
\begin{align}
	\forall x, y \in P \quad \exists z \in W & \quad : z = \Psi(x,y)\\
                                             & \quad b(z) = w(x,y) 
\end{align}
\end{definition}

\begin{definition}
The function $\Psi: \mathbb{X} \times \mathbb{X} \to W$ is called the \textbf{pairwise mapping function} and projects the pair of points from the original search space into a point in the pairwise training dataset.
\end{definition}

There are several possible pairwise mapping functions $\Psi$. The simplest one just combines the components of the two solutions. For instance, if the optimisation problem is defined by $n$ real numbers $\mathbb{X}=\mathbb{R}^n$, this function is $\Psi: \mathbb{R}^n \times \mathbb{R}^n \to \mathbb{R}^{2n}$:

\begin{equation}\label{eq:mapping1}
     \Psi(\vec{x}, \vec{y}) = (\overbrace{x_1, x_2, \dots, x_n}^{\vec{x}}, \overbrace{y_1, y_2, \dots, y_n}^{\vec{y}}) 
\end{equation}

This study has used an extended mapping function $\Psi^+$ that also includes the difference vector between the two solutions  $\Psi^+: \mathbb{R}^n \times \mathbb{R}^n \to \mathbb{R}^{3n}$:

\begin{equation}\label{eq:mapping2}
\begin{split}
     \Psi^+(\vec{x}, \vec{y}) = (\overbrace{x_1, x_2, \dots, x_n}^{\vec{x}}, \overbrace{y_1, y_2, \dots, y_n}^{\vec{y}},
     \\
     \overbrace{x_1-y_1, x_2-y_2, \dots, x_n-y_n}^{\vec{x}-\vec{y}}) 
\end{split}
\end{equation}

Experiments have shown that this extra information can enhance the performance of the surrogate model in some configurations of the problem and/or the surrogate. For example, if the surrogate cannot model the distance internally, such as a DT, this extension makes this information available for making decisions. 
 
\subsection{Machine Learning Models for Pairwise Surrogates}
Although there is the possibility of choosing a different set of binary classifiers as the algorithm to train the pairwise surrogate models, this study proposes to use the classifier (discrete) version of the regression algorithms mentioned in Section~\ref{sec:ml}.

Table~\ref{tab:surface-to-pairwise} shows the regression algorithms used to construct the \textit{surface surrogate} and the equivalent classification algorithm used to model the \textit{pairwise surrogate} equivalents.

\begin{table}[!htp]
    \centering
    \begin{tabular}{ll}
        \hline
        \textbf{Surface (regression)} & \textbf{Pairwise (classification)}\\
        \hline
        Ridge/R   & Ridge/C    \\
        RF/R      & RF/C \\
        XGBoost/R & XGBoost/C  \\
        DT/R      & DT/C                   \\
        MLP/R     & MLP/C\\
        \hline    
    \end{tabular}
    \caption{Surface and pairwise surrogate algorithms.}
    \label{tab:surface-to-pairwise}
\end{table}

This selection of machine learning algorithms covers the most representative methods. Ridge, which is mainly a regressor, also has an implementation to use it as a classifier and exemplifies the possibility of using native regression models as pairwise surrogates. On the other hand, decision trees (DT) are intended to be classifiers, although there exist regression variants that use simple linear models at their leaves. The objective of this study is not to find the best machine learning model for the surrogate but analysing the general performance in conjunction with the different surrogate strategies (see Section~\ref{sec:strategies}). 

\section{Experimental Scenario}\label{sec:experiments}
For the experimental analysis of the different surrogate models, we have selected a well-known benchmark proposed in \citep{lozano2011editorial} for the SOCO'2011 special issue on the ``\textit{Scalability of evolutionary algorithms and other metaheuristics for large-scale continuous optimisation problems}''.  In order to adjust the experimentation to the inherent characteristics of the problems that are normally solved with the help of surrogate models, we have conducted the following changes to the usual experimental conditions of this benchmark: each configuration is run 15 times instead of the usual 25, same run number is guaranteed to start from the same initial population, maximum fitness evaluations has been fixed to 15D and problem size to 
$D=50$ (further details can be found in \ref{app:experimental-scenario})

The primary goal of this article is to analyse the potential of using surrogate models, based on different strategies, in order to save computation time while achieving a good-quality solution. It is not expected for any of the configurations to reach the optimal value in just 750 quality function evaluations, but to provide the best possible solution given the aforementioned budget. Additionally, this study aims to compare the pairwise surrogate modelling techniques (based on classifiers) with the surface (regression-based) equivalents. In order to face different models and strategies among them, the lowest quality value reached in the optimisation will be used as the comparison criterion. 

\subsection{Reference Algorithm}\label{sec:de}
Several advanced self-adaptive hybrid algorithms have dominated continuous optimisation competitions during the last years \citep{tvrdik2013competitive,latorre2013large,guo2014enhancing, qin2013differential,tanabe2013evaluating,latorre2015comprehensive,draa2015sinusoidal,latorre2017comparison}. Furthermore, many of these approaches have also exhibited a great performance in real-world problems beyond these competitions \citep{molina2019comparing}. Many of these approaches have in common the usage of some version of Differential Evolution (DE) as the core algorithm. However, due to the nature of the problems under consideration (computationally complex problems and small budget of quality function evaluations), and the characteristics of those algorithms (hybrid and/or self-adaptive, i. e., they may need some warm-up time), we have decided to use the simpler yet effective basic DE as the optimiser for our experiments. Further details about the specific implementation can be found in \ref{app:reference-algorithm}.

\subsection{Statistical evaluation}\label{sec:evaluation}

To assess the relevance of the results obtained in our experiments, we have conducted a statistical validation based on the following procedure. First, the average ranking of the algorithms on the whole set of functions has been computed. Then, the algorithm with the best overall ranking has been selected as the reference algorithm. Finally, pairwise comparisons have been conducted between the reference algorithm and all the other methods with both the Friedman and the Wilcoxon tests. Furthermore, p-values have been corrected to account for the family-wise error in multiple comparisons with the Holm method.

\subsection{Hyper-parameter definition}
Although this study is meant to be a general comparison of different machine learning models, approaches, and strategies in the context of an integrated optimisation framework, there are several parameters of such framework that need to be adjusted in order to obtain reasonable results. Whenever it has been possible (in terms of computational cost) the parameters have been independently.

In this section we analyse each of the parameters identified in the integrated framework. Among them, two of them are of paramount importance: the \textit{warm-up cycles} and the \textit{retrain frequency} of the models. Additionally,  several specific parameters that control either early convergence detection or other computational aspects are also considered in the following sections. 

\subsubsection{Retrain frequency}
The \textit{retrain frequency} establishes when the framework updates the surrogate model with the available data. For this study, we have set this frequency to every generation. Hence, the machine learning model is re-trained at the end of every DE generation (regardless of whether the framework actually evaluates the quality function or not). This value is a reasonable trade-off, considering that training a machine learning model for a few hundred data points is by far more efficient than the execution of a quality function that might involve a simulation or a similar computation.

\subsubsection{Pairwise approaches}
\label{sec:pairwise_approaches}
One of the main characteristics of the \textit{pairwise surrogate} models is that the available data points can grow very fast (quadratically with the number of quality function evaluations). For each new individual, multiple new data points can be generated by comparing it with all the previously created individuals. Although this feature can boost the initial training capability of the model (it can train on a larger dataset with few new individuals), in the long run it might be an efficiency problem when updating the model (depending on the \textit{retrain frequency} mentioned before), especially as the optimisation progresses.

To deal with this problem, we have limited the pairwise combinations included in the training data set. The authors are fully aware that it somehow limits one of the main characteristics of the \textit{pairwise surrogates}. Nonetheless, keeping the same \textit{retrain frequency} seems fairer when comparing the two main surrogate strategies. To instrument the same limit for all the \textit{pairwise surrogate} models, the study defines a \textit{trail size}, that limits how many results are combined with each new result to create pairs of data points for the training set. The experiments have set this \textit{trail size} parameter to the 45 most recent results.

\subsubsection{Warm-up Phase}
The third parameter establishes the required number of samples the surrogate needs to create the initial model. These experiments calibrate this \textit{warm-up} value for each proposed machine learning algorithm, because this number dramatically depends on the characteristics of the algorithm and how much information it requires to make a reasonably accurate first prediction. The selected value is based on the ranking of the 19 functions according to the same evaluation criteria mentioned in Section~\ref{sec:evaluation}.

Table~\ref{table:warmUpValues} summarises the final configuration used for each machine learning algorithm and surrogate approach. More information about this process can be found in \ref{app:warm-up}.

\begin{table}[hbp]
	\centering
	\begin{tabular}{lcc}
		\hline
        & Regressor (/R) & Classifier (/C)  \\
        \hline
        RF      & 40             & 20        \\
        MLP     & 30             & 2         \\
        Ridge   & 10             & 2         \\
        DT      & 30             & 4         \\
        XGBoost & 2              & 6         \\
		\hline
	\end{tabular}
    \caption{Number of generations in the Warm-up phase.} 
    \label{table:warmUpValues}
\end{table}

\subsubsection{No-improvement condition}
\label{sec:noimprovementcondition}
Stagnation (no progress during the search) might happen when using surrogate models when they underestimate the quality of the candidate solutions. In those cases, the combination of the surrogate strategy and the optimisation search is biased in a way that prevents any candidate solution from meeting the conditions for being evaluated with the actual quality function. This effect is an artefact that causes the optimisation to get stuck and iterate many generations without spending the quality function evaluation budget. To avoid this undesired effect, the optimisation algorithm must detect this situation and stop the search. We have included an additional convergence criterion based on a parameter that establishes the \textit{maximum number of generations without improvement}.

\section{Analysis of the Experimental Results}\label{sec:results}
This section presents the results of the experiments conducted on the benchmarks introduced in Section~\ref{sec:experiments} according to the statistical analysis methodology and the hyper-parameters calibration carried out, also presented in the same section.

The first part of this section compares the default strategy implementation for the surrogate models when using different machine learning algorithms. This preliminary analysis also compares the surface approach with the \textit{pairwise} surrogate model (introduced in Section~\ref{sec:pairwise}). 

The second part of the section carries out an analysis of the different strategies (see Section~\ref{sec:strategies}). This analysis compares the strategies independently and in combination with all the algorithms and models (surface and pairwise). 

\subsection{Default Strategy}

Tables~\ref{table:rankingBaselineAll} compare the ranking for both the surface and the pairwise surrogate models. The results clearly show that the DT/C strategy not only outperforms all the other pairwise approaches, but also all their surface counterparts. All the differences are statistically significant. Furhter detailed tables with all the results have been included as supplementary material in the \ref{app:sec:default_strategy} section. 

\begin{table}[!htp]
	\centering
	\begin{tabular}{lc}
		\hline
		& Ranking \\ 
		\hline
		DT/C & 1.16 \\ 
		XGBoost/R & 4.26 \\ 
		MLP/C & 4.37 \\ 
		Ridge/R & 4.68 \\ 
		Ridge/C & 5.21 \\ 
		XGBoost/C & 6.79 \\ 
		DE & 7.32 \\ 
		MLP/R & 7.32 \\ 
		DT/R & 7.42 \\ 
		RF/C & 8.32 \\ 
		RF/R & 9.16 \\ 
		\hline
	\end{tabular}
    \caption{Average Ranking for the surface and pairwise approaches without improvement strategies.} 
    \label{table:rankingBaselineAll}
\end{table}

\subsection{Analysis of the Different Improved Strategies}\label{sec:strategies-analysis}
This second part of the experimentation covers the different alternative strategies presented in Section~\ref{sec:strategies}. This analysis takes each of these strategies (Default Strategy, Probability-based Strategy (\textit{Prob}), Quality Distance Strategy (\textit{Qual}), and Candidate Diversity Strategy (\textit{Diver})) and compares them independently and in combination. This type of analysis compares not only the individual contribution of each strategy but also the synergies resulting from their combination.  

As this experimentation includes many comparative analyses carried out for each of the strategies combined with the machine learning algorithms and for both the surface and the pairwise surrogate models, the detailed tables with all the results have been included as supplementary material in the \ref{app:sec:strategies-analysis} section at the end of the article. This section summarises these results, grouping them according to the corresponding surrogate model used: surface vs pairwise approaches. Finally, at the end of this section we provide a comparative analysis of the best overall strategies. As stated before, the reader is referred to the \ref{app:sec:strategies-analysis} section for more detailed information. 

\subsubsection{Surface Surrogate Results}

\begin{figure*}[!htp]
    
    \centering
	\includegraphics[width = 0.40\textwidth, keepaspectratio=true]{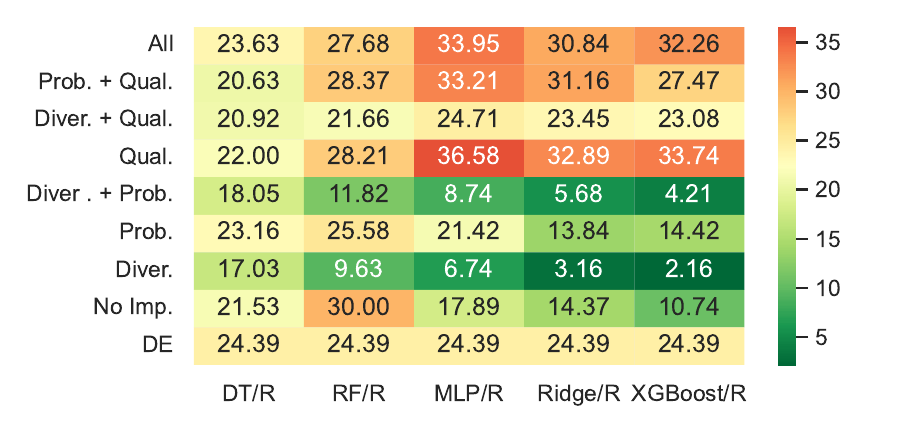}
	\includegraphics[width = 0.40\textwidth, keepaspectratio=true]{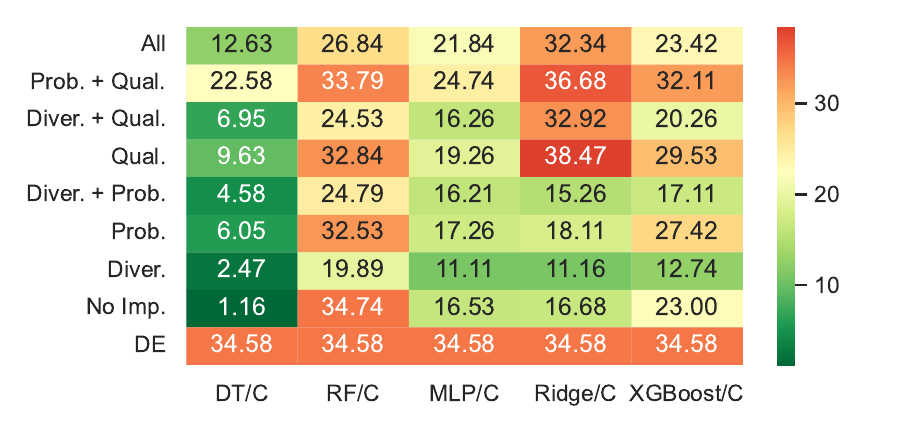}
	\caption{Heat maps of the rankings for each combination of model and strategy for the surface approach (left) and pairwise approach (right).}
	\label{fig:HeatMapSurface}
\end{figure*}

This section reports the results of the different strategies (and combinations) of DT/R, MLP/R, Ridge/R, and XGBoost/R models. Figure~\ref{fig:HeatMapSurface} depicts, on the left, the relative ranking of all the combinations of machine learning algorithms and strategies (6 machine learning algorithms plus the baseline DE without surrogates times 8 strategies).

Among all the tested combinations, the best performing ones are those using the Candidate Diversity Strategy (\textit{Diver}). The statistical analysis, in this case, reveals significant differences between XGBoost/R+Diver. and the best performing strategies for the other machine learning algorithms. The only exception is Ridge/R+Diver, which is a rather simple machine learning model, but not performing much worse than the reference algorithm (and only for the Friedman test).

As shown in these plots, the strategy seems to be more relevant than the algorithm. Moreover, the use of the Candidate Diversity strategy, in combination with any other strategy, always obtains competitive results. Nonetheless, the best overall results are obtained when this strategy is used standalone.

\subsubsection{Pairwise Surrogate Results}
Similarly to what was shown in the case of the Surface Surrogate Models, this section presents the results for the pairwise models based on the RF/C, DT/C, MLP/C, Ridge/C, and XGBoost/C machine learning models, respectively. Figure~\ref{fig:HeatMapSurface} presents on the right, analogously to what was shown for the surface models, a heat map with the relative rankings of all the Pairwise Models.

A quick analysis of these results reveals that the Candidate Diversity Strategy (\textit{Diver}) is, again, the best performing one for most of the machine learning algorithms, as it was for the Surface Surrogate Models. The only exception here is the DT/C model (without any strategy) which, surprisingly, obtained the best ranking. Results also report significant differences for the statistical validation between DT/C, the reference algorithm (best ranking) and all the other combinations of machine learning algorithms and strategies.

Unlike what happened with the surface approaches, the most influential component is, in this case, the use of DT/C as the machine learning model. For the remaining cases, the Candidate Diversity Strategy is generally better than any other configuration, except for DT/C for which, the simpler the configuration, the better are the results obtained.

\subsubsection{Pairwise vs Surface Surrogate Results}\label{sec:final-comparison}

This final analysis compares the best combination of machine learning model and strategy (\textit{Diver} in most of the cases) for both the surface and the pairwise models. The ranking of this overall comparison, shown in Table \ref{table:rankingAllImpBoth}, indicates that the best surrogate configuration is DT/C. By just observing the ranking values, it seems clear that there is a second group of algorithms, including XGBoost/R+Diver. and Ridge/R+Diver., with a similar performance but worse than DT/C. The third group of algorithms includes Ridge/C+Diver, MLP/C+Diver, MLP/R+Diver, and XGBoost/C+Diver, all of them with very similar rankings. At the bottom of the ranking we can find RF/C+Diver, RF/R+Diver. and DT/R+Diver. (being the last one slightly worse than the other two). Finally, the baseline algorithm, the DE with no surrogate model, is the worst alternative.

This analysis reveals significant differences in the statistical comparison between the reference algorithm (DT/C) and all the other approaches for both statistical tests, with the only exceptions of Ridge/R+Diver. and XGBoost/R+Diver. in the Friedman test. (See \ref{app:sec:final-comparison} in the Supplementary Material)

\begin{table}[!htp]
\centering
\begin{tabular}{lc}
  \hline
 & Ranking \\ 
  \hline
DT/C & 1.16 \\ 
  XGBoost/R + Diver. & 2.21 \\ 
  Ridge/R + Diver. & 2.89 \\ 
  Ridge/C + Diver. & 4.95 \\ 
  MLP/C + Diver. & 5.32 \\ 
  MLP/R + Diver. & 5.89 \\ 
  XGBoost/C + Diver. & 5.95 \\ 
  RF/C + Diver. & 8.16 \\ 
  RF/R + Diver. & 8.61 \\ 
  DT/R + Diver. & 9.92 \\ 
  Benchmark & 10.95 \\ 
   \hline
\end{tabular}
\caption{Average ranking for the best strategy for every surrogate-approach combination.} 
\label{table:rankingAllImpBoth}
\end{table}

\subsubsection{Best approaches vs Kriging results}\label{sec:kriging-final-comparison}

Additionally, two Kriging models, the most common surrogate approach, are included for a final comparison as both online (trained during the optimisation) and offline (training before optimisation and replacing the objective function) models. Both Kriging approaches use a Gaussian Process Regressor (GPR) with the default scikit-learn configuration, which uses a Radial Basis Function (RBF) kernel. The online model is an alternative to the other machine learning algorithms, whereas the offline Kriging surrogate is trained by sampling the space with 749 candidate solutions (selected by a Latin Hypercupe), and saving the remaining sample to real-evaluate the final proposed solution. As the Kriging model replaces the original function, budget limitation is waved for offline surrogate model. To reach the proposed solution, the same DE configuration will be executed with a budget limit of 3M (60,000 executions per dimension), enough to consider that optimisation is not early-ended.

Table \ref{table:rankingKriging} shows the comparison between the best alternatives found in the previous ranking vs the two mentioned Kriging approaches. The reference algorithm, DT/C, have shown significant differences against both Kriging models. (See \ref{app:sec:kriging-comparison} in the Supplementary Material)

\begin{table}[!htp]
\centering
\begin{tabular}{lc}
  \hline
 & Ranking \\ 
  \hline
DT/C & 1.63 \\ 
  XGBoost/R + Diver. & 2.89 \\ 
  Kriging (Online) & 3.11 \\ 
  Ridge/R + Diver. & 3.89 \\ 
  Kriging (Offline) & 3.89 \\ 
  Benchmark & 5.58 \\ 
   \hline
\end{tabular}
\caption{Average ranking for the found solutions vs Kriging models} 
\label{table:rankingKriging}
\end{table}

\section{Discussion}\label{sec:discussion}

There are several interesting aspects to discuss concerning: (1)~the performance of different algorithms, in particular, DT/C which ranks the highest in the experiments, (2)~the relationship between the performance of the surrogate models and their accuracy as supervised learners, (3)~the evaluation of the computational efficiency in terms of resources reduced by the use of surrogates, and (4)~the analysis of the surrogate model contribution during the course of the optimisation. 

\subsection{DT/C Performance Analysis}

Table~\ref{table:rankingAllImpBoth} shows that DT/C is the surrogate configuration with the best performance and a remarkable difference in ranking with the rest. This is confirmed by the statistical validation, in which results are significant compared to all the other alternatives (according to the Wilcoxon test), or most of them (if the Friedman test is considered). This result is interesting from several points of view:
\begin{itemize}
\item DT/R variants using the surface surrogate model perform significantly the worst (being DT/R+Diver. the worst of the strategies, in the final analysis, see Table~\ref{table:rankingAllImpBoth}).

\item DT/C is the only configuration included in the final analysis not using any of the additional strategies. The rest of the configurations get a significant improvement when using the Candidate Diversity Strategy, but not in the case of DT/C.
\item DT is not a particularly powerful machine learning algorithm in the wide variety of supervised learning models. It can capture more complex patterns than a regression model (e.g., Ridge) but it does not have the potential of more advanced models such as XGBoost or MLP. For instance, the performance of RF is expected to be better than DT. 
\end{itemize}

There are certain explanations for some of these questions. For instance, DT was originally designed to solve discrete classification problems, and DT/R is an extension of this original approach to deal with continuous regression problems. In this sense, DT/C is a more natural matching between the technique and the problem than DT/R. Since the way the surrogate training is tackled affects its performance, the pairwise approach opens new options to build surrogate models that can improve its performance by matching the problem approach seen from the model viewpoint to its original design. However, this does not seem to tell the whole story. 

\begin{figure*}[!htp]
    \centering
	\includegraphics[width=0.8\textwidth, keepaspectratio=true]{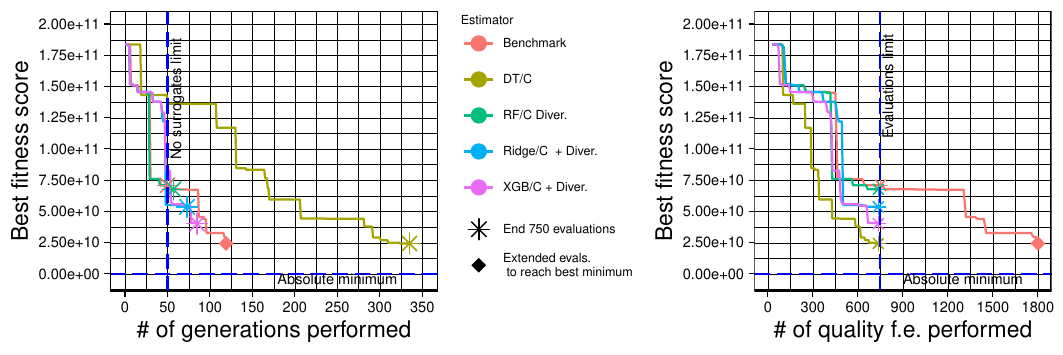}
	\caption{Example of the optimisation result per generation (left) or quality function evaluations, f.e. (right). Note: DE is an extended version of the baseline algorithm (with no surrogates) that has been run to reach the same result than the best performing configuration with a surrogate (DT/C).}
	\label{fig:DTOutperformance}
\end{figure*}

When analysing the performance of a sample run for a given problem (see Figure~\ref{fig:DTOutperformance}), we can see that DT/C manages the budget (of quality function executions) rather efficiently. In this plot, the DE execution has been extended to 110 generations (1650 quality function evaluations compared to the 750 for the surrogate models) to find the point at which it reaches the same quality value that DT/C obtained. The left-hand side plot of Figure~\ref{fig:DTOutperformance} shows the results by generations. From this plot, it is easy to see how, while the rest of the algorithms do not reach the 100 generations before running out of quality function evaluations, DT/C reaches more than 300 generations. However, at almost any previous generation, DT/C is getting a worse value per generation. This per-generation plot also allows analysing that RF/C (which has the most prolonged warm-up phase) is mostly paired with the DE without surrogates, but it does not progress for more than 50 generations, which is an indication that it is not performing any significant filtering. 

Furthermore, the right-hand side plot of Figure~\ref{fig:DTOutperformance} displays the same results, but instead of showing them by generations performed, it is plotted using the number of quality function evaluations performed in the X-axis. This representation is almost the inverse of the one mentioned above. Looking at the improvement per function evaluation, DT/C is getting the best result almost consistently. Furthermore, we can rank the performance starting from the bottom (better quality) and moving up using this representation. Hence, the order in performance for this particular example is first DT/C, then XGBoost/C+Diver, Ridge/C + Diver, and finally RF/C+Diver. This ranking is not always necessarily the same, but it is a relatively common scenario. 

Therefore, as a conclusion of this first part of the discussion, it is likely to consider that encouraging the search algorithm to run for more generations generally improves the quality of the solution. It is necessary to exclude from this conclusion those cases when the surrogate underestimates the quality of almost all samples and, therefore, optimisation gets stuck (see Section~\ref{sec:noimprovementcondition}).

\subsection{Accuracy of the Supervised Learners Compared to Their Performance}
In essence, the use of surrogates is tightly aligned with the ability of the machine learning algorithm to model the quality function. This interpretation is particularly relevant in those approaches using the surrogate function as a substitute for the quality function, using the search algorithm to optimise the surrogate model instead of the actual function. Nevertheless, this consideration poses the question of analysing how accurate the surrogates models are in relation with their performance. This part of the analysis has been conducted using both surrogate models (Pairwise vs Surface) as well as the different machine learning models and their best-performing strategies. 

For these experiments, accuracy has been measured according to the actual purpose of the surrogate in the optimisation process. This is, in the particular case of the DE algorithm used in this study, to determine whether the new candidate solution is better or not than the current one. This interpretation works for both the surface surrogate model (which predicts the continuous value of the quality function to compare this prediction with the actual value of the current solution), as well as for the pairwise surrogate model (which directly predicts the result of this comparison). 

Specifically for this analysis, the results come from always computing the actual quality function values, but conducting the search according to the surrogate model plus the strategies, for each configuration as in the previous experiments. Hence, just in this case, we always have the prediction of the model and the actual results from the execution of the quality function. To interpret these results, we use the following notation: a \textit{positive case} is when the candidate solution \textit{is better than} the current solution ($w(x, x')=1$ in Equation~\ref{eq:pairwise-base} for $x$ being the current solution and $x'$ being the candidate solution). The evaluation of the surrogates is then calculated with three well-known supervised learning metrics:
\begin{itemize}
\item \textit{Accuracy}: The ratio of \textit{true positive} plus \textit{true negative} cases compared to all the cases (candidates generated by the optimisation algorithm).
\item  \textit{Sensitivity}: The ratio of \textit{true positive} cases compared to all the cases that are positive. This metric measures how good the model is to detect promising candidates.
\item  \textit{Specificity}: Similarly, the ratio of \textit{true negative} cases compared to all the negative cases. This metric measures how good the model is to filter out those cases that do not improve the solution. 
\end{itemize}

\begin{figure*}[!htp]
\centering
	\includegraphics[width=0.8\textwidth, keepaspectratio=true]{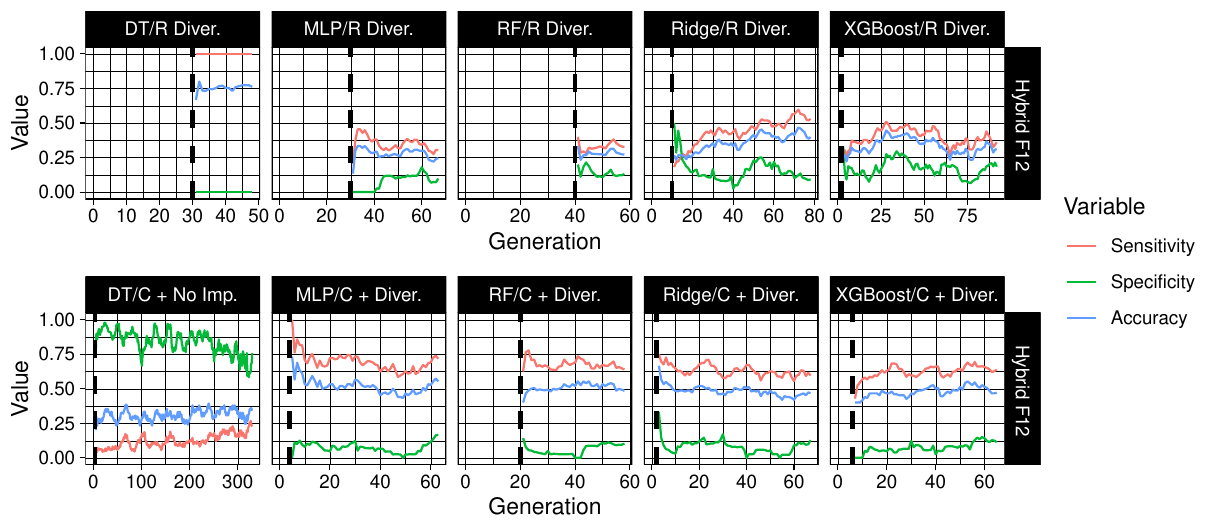}
	\caption{Sensitivity, specificity and accuracy for the Hybrid F12 function with the best improvement. Vertical dashed lines define the point when warm up ends.}
	\label{fig:sensHybridF13}
\end{figure*}

Figure~\ref{fig:sensHybridF13} show these metrics for Hybrid F12, for each of the best models included in the final comparison. DT/R+Diver. is predicting all the cases as positives, not behaving differently than the baseline DE. However, in terms of accuracy, it has the highest value (most of the candidate solutions in the first generations are indeed improving the corresponding current solutions). If we discard this degenerate case, the best accuracies correspond to the pairwise surrogate models, typically in the 0.5 threshold (compared to the 0.25 averaged by the surface-based surrogates). Surface surrogate models tend to have very similar sensitivity and specificity values, but low accuracy. These results mean that the errors they are committing are rather balanced between over and underestimation of the candidate solutions. This behaviour is also expected because they are trained to model the problem as a continuous function. \ref{app:sec:all-accuracy} includes all the accuracy analysis figures.

The most relevant insight obtained from this analysis is that, whereas most of the algorithms have a higher sensitivity than specificity, for DT/C we have the opposite case. DT/C excels in identifying those non-interesting solutions and filtering them out, although many promising solutions are also discarded. This conservative behaviour (in a problem which, at least in its first generations, includes many cases whose candidates are better than the corresponding current solutions) pulls down the accuracy of the DT/C to lower values than the other pairwise surrogate models. However, this is not translated into a worse performance in terms of optimisation results, quite the contrary. 

\subsection{Computational Resource Usage}
\label{sec:computationaResourceUsage}

The whole rationale behind the use of surrogate models is to have an approximate function that allows the optimisation algorithm to estimate the quality of new candidates and (at least in those alternating between surrogates and the actual quality function) decide whether to calculate the actual quality function or discard the candidate. 

\begin{definition}
For the measurement of the computational efficiency, this analysis includes a \textbf{computational cost reduction} metric $\Delta^n e$ to calculate the difference in the number of quality function evaluations required to find the same value by the baseline DE algorithm:  
\begin{equation}
	\Delta e_n = m - n \quad : q(x_s^n) = q(x_b^m)
\end{equation}
Let $x_s^n$ be the best solution obtained by the surrogate model $s$ after $n$ quality function evaluations, and $x_b^m$ be the best solution found by the baseline DE algorithm $b$ after $m$ quality function evaluations.
\end{definition}

This metric $\Delta e_n$ measures how efficient the surrogate is to find the best solution compared with the baseline algorithm without surrogates when the surrogate model has used $n$ quality functions. As the analysis conducts the experimentation under the conditions mentioned in Section~\ref{sec:experiments}, the most interesting value of this metric is the one obtained at the end of the optimisation.

\begin{equation}
	\Delta e = \Delta e_{15D} = \Delta e_{750}
\end{equation}

\begin{figure}[!htp]
\centering
	\includegraphics[width=0.7\columnwidth, keepaspectratio=true]{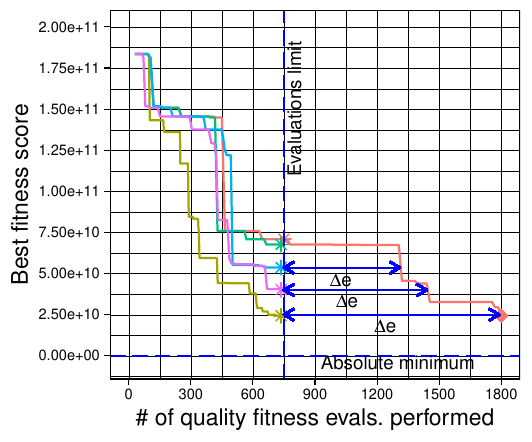}
	\caption{Example of the computational cost reduction calculation.}
	\label{fig:SavingExample}
\end{figure}

Figure~\ref{fig:SavingExample} graphically depicts an example of how this metric is calculated.

\begin{definition}
The \textbf{relative computational cost ratio} $\delta e_n$ is  calculated as the fraction of the number of quality function evaluations required by the surrogate configuration divided by the number of quality function evaluations consumed by the DE algorithm to find the same value:  

\begin{equation}
	\delta e_n = \frac{n}{n + \Delta e_n} = \frac{n}{n + m - n} = \frac{n}{m}
\end{equation}
\end{definition}

The range of values of this $\delta e_n$ metric varies from $0$ to, potentially $\infty$. A value of less than $1$ means that the surrogate consumed fewer quality function evaluations than the DE. However, the DE algorithm might also generate a good-quality candidate early in the optimisation process that is discarded by the surrogate. In the few cases this happens, the final result delivered by the surrogate model would be worse or equal than DE solution fitness value. In those cases, the $\delta e_n$ will be greater than $1$, which significantly skews the plot. Similarly to the case of the computational cost reduction: 

\begin{equation}
	\delta e = \delta e_{15D} = \delta e_{750}
\end{equation}

\begin{figure*}[!htp]
\centering
	\includegraphics[width=0.8\textwidth, keepaspectratio=true]{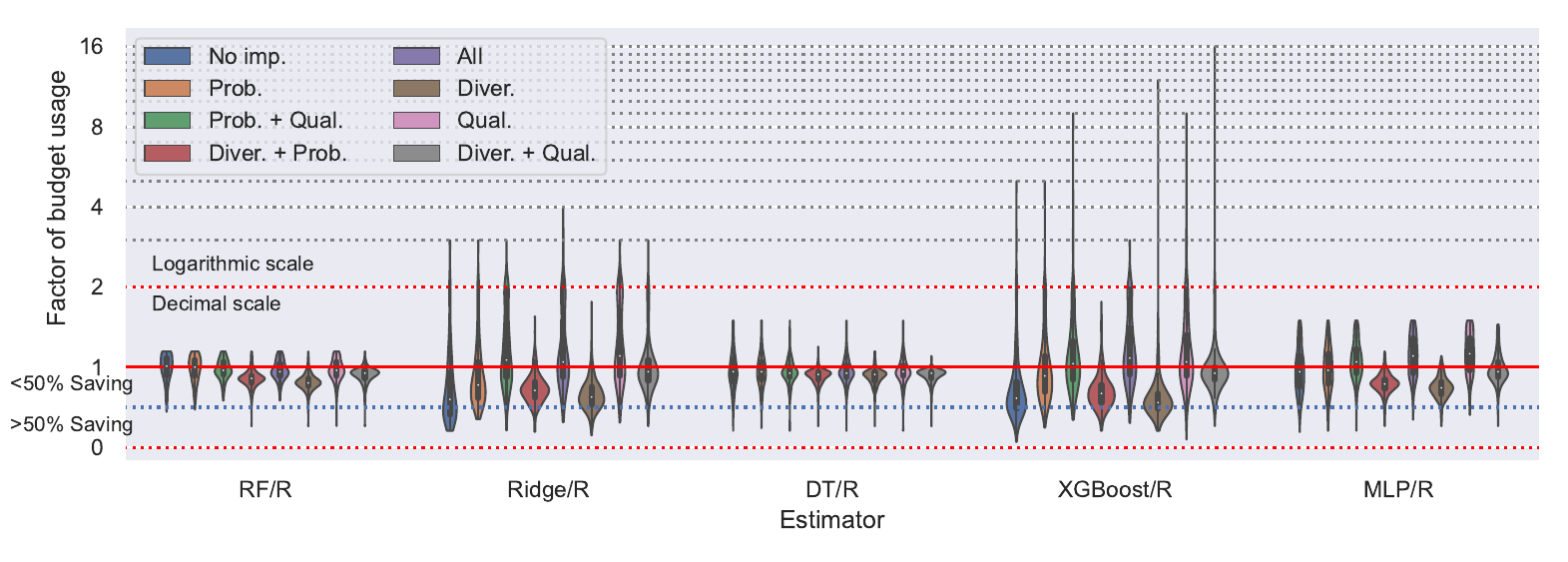}
	\caption{Distribution of the relative computational cost ratio $\delta e$ for every algorithm and strategy using surface surrogate models.}
	\label{fig:computationalSavingSurface}
\end{figure*}

\begin{figure*}[!htp]
\centering
	\includegraphics[width=0.8\textwidth, keepaspectratio=true]{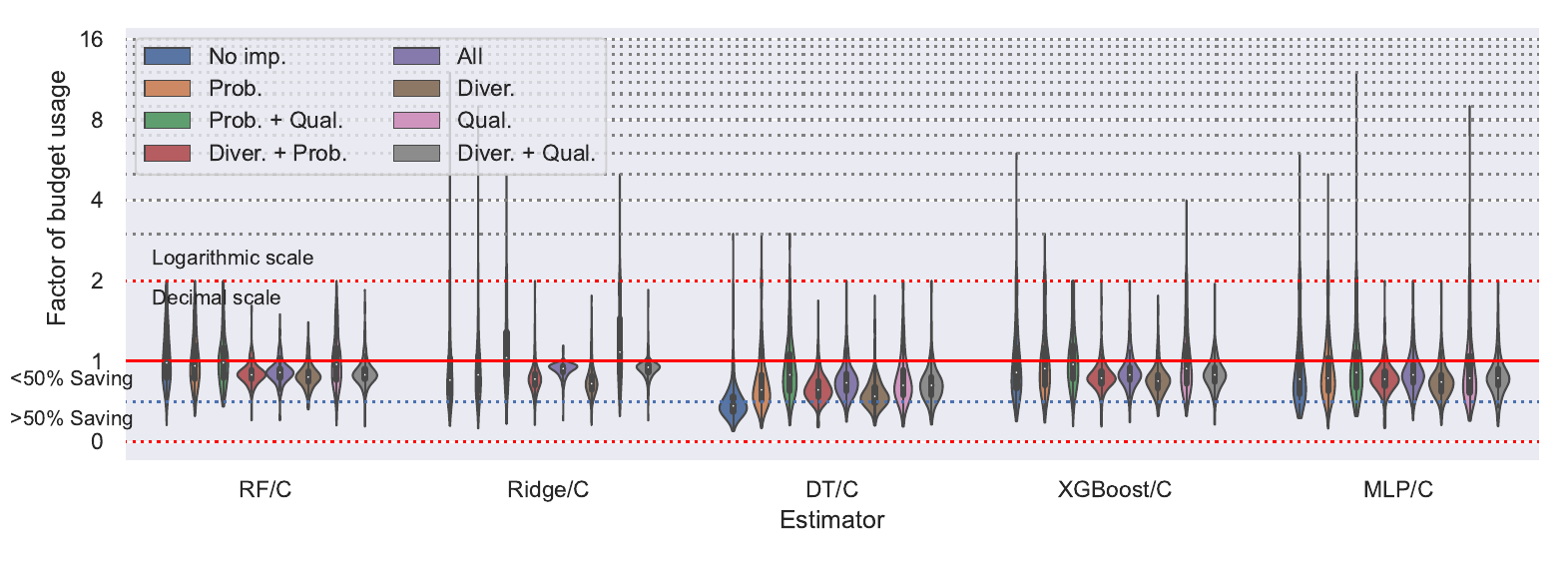}
	\caption{Distribution of the relative computational cost ratio $\delta e$ for every algorithm and strategy using pairwise surrogate models.}
	\label{fig:computationalSavingPairwise}
\end{figure*}

Figure~\ref{fig:computationalSavingSurface} shows the distribution of $\delta e_n$ for the surface surrogate model configurations, in the format of violin plots, for each machine learning model and for all of the surrogate strategies. Similarly, Figure~\ref{fig:computationalSavingPairwise} displays the equivalent plots for pairwise surrogate models. The reference value in these plots (in solid red) is $\delta e=1$, for which the surrogate is not able to reduce the computational cost of the baseline DE. Below this line, the dotted blue line represents the $50\%$ saving threshold ($\delta e=0.5$). This means that the lower the distribution is, the better. Values greater than $1$ represent worse performance compared to the baseline DE, and for those values greater than $2$, the plot uses a logarithmic scale. 

The analysis of these plots indicates that, in general, small relative computational cost ratio $\delta e$ correlates with the configuration ranking. Nonetheless, in the case of XGBoost/R and Ridge/R configurations, which were the best performing surface surrogate models, they exhibit long tales in which these algorithms seldom skipped good quality solutions early in the optimisation that happened to be a better result than the final value obtained by the surrogate-based optimisation. This characteristic represents how aggressive the surrogate model is. This aggressiveness in performing the candidate filtering tends to produce better results on average (check the body of the violins and the mean point and variance boxes) but they risk being less reliable. The two worst performing configurations (based on RF/R and DT/R) display low variance plots with reduced variability. 

On the contrary, the pairwise surrogate models (Figure~\ref{fig:computationalSavingPairwise}) show that the highest-ranked configurations (based on DT/C) have lower average values in their relative computational cost ratio $\delta e$ but keep the variability of the surrogate-based optimisation algorithm much lower than in the case of the best surface surrogate models. This result implies that the excellent performance achieved by DT/C is mainly based on a very conservative behaviour in the filter, that consistently allows it to run the optimisation algorithm for more generations, rather than a fine-grain prediction of the quality function.

\subsection{Surrogate Contribution Evolution}
\label{sec:surrogateContributionEvolution}

The search process of an optimisation algorithm is well known to be characterised by different phases. In general, the main distinction comes from exploration vs exploitation, as stated in \citep{chen2009optimal} among other authors.  For this study, the experiments have run with a reduced budget, in terms of quality function evaluations (hence, limited to the early exploration phase). However, there is a possibility to identify phases in the search process, in which the behaviour of the surrogate model is different. Nevertheless, these behaviours may depend on the problem to optimise.

\begin{definition}
Let $\zeta_d(i)$ be the \textbf{mean improvement per budget unit} over the last $d$ generations, defined as:
\begin{equation}
	\zeta_d(i) = \log{\left(\frac{|q(x^{v(i)}) - q(x^{v(i-d)}|}{v(i)-v(i-d)}\right)}
\end{equation}
Where $v(i)$ is the number of quality function evaluations consumed at generation $i$, $x^{v(i)}$ is the best solution found after $i$ generations, and $q(x^{v(i)})$ is the quality function value of that solution.
\end{definition}

The value of $\zeta_d(i)$ gives a measure of the performance of the optimisation process per generation $i$. This value represents the mean improvement per fitness function evaluation in a pre-defined sliding window of size $d$ generations. Later, it is adjusted to logarithmic scale to reduce the bias produced by first generations making the majority of the improvements. This metric is highly dependant on the problem features so it is not recommended unless comparisons are made against the same problem. When this metric applies, it depicts the average efficiency of a single fitness evaluation, being the higher the value, the better. It is also remarkable that a high efficiency value would usually be related to a higher number of generations as it would mean that the surrogate model is doing its job properly and only allowing the execution of useful proposals. 

\begin{figure*}[!t]
    \centering
	\includegraphics[width=0.7\textwidth, keepaspectratio=true]{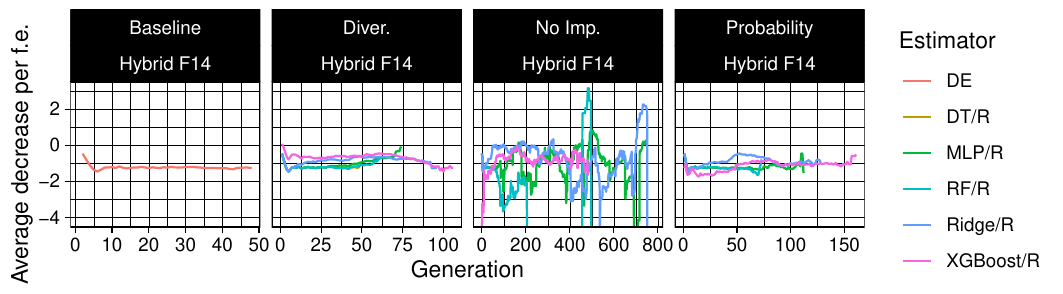} \\
	\includegraphics[width=0.7\textwidth,keepaspectratio=true]{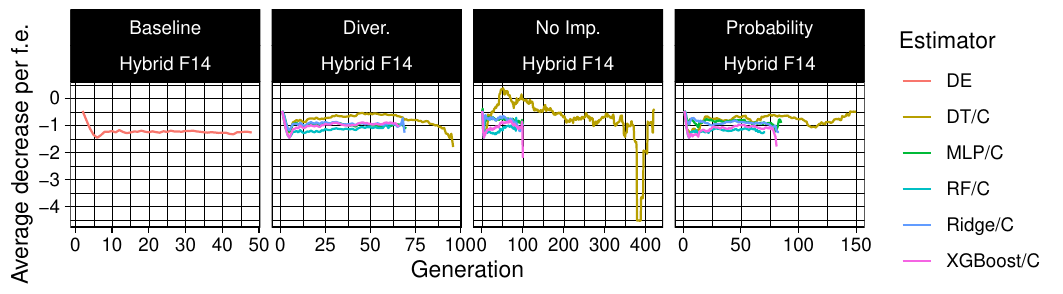}
	\caption{Mean decrease per budget unit (fitness evaluation, f.e.) in the last 40 DE generations.}
	\label{fig:chunks}
\end{figure*}

Figure \ref{fig:chunks} shows $\zeta_{40}(i)$ values for two sample functions. This measure implements a sliding window of 40 generations back (limited to generation 0 if less than 40) to average the results for a significant number of values to have some smooth form. The peaks appearing during certain moments of the optimisation process suggest phases in which one algorithm outperforms the others. This result encourages to make a further study in using several surrogates leveraging them during the different phases.

\section{Performance Analysis in a Real-World Simulation Application}
\label{sec:realworld}

The results of the previous section, which involved benchmark functions whose computation time is relatively short, allowed us to perform a complete comparative analysis. However, surrogates address problems that involve much more computationally expensive functions, mostly simulation-based optimisation problems. With the benchmark functions, it is significantly faster to execute the function rather than evaluate the surrogates. 

In order to analyse the performance based on time, we propose an analysis based on a real-world simulation problem. For this new case, the computational time is much more, and there is a clear benefit in using surrogates to filter uninteresting solutions. The problem to optimise was proposed in GECCO 2021 Industrial Challenge under the name \textit{``Optimisation of a simulation model for a capacity and resource planning task for hospitals under special consideration of the COVID-19 pandemic''}\footnote{https://www.th-koeln.de/informatik-und-ingenieurwissenschaften/gecco-2021-industrial-challenge-call-for-participation\_82086.php}. Further details about the participation of the proposed method in the competition can be found in \citep{naharro2021gecco}. The problem is a black-box optimisation function wrapped in a Docker image \citep{Bartz-Beielstein2020}, which receives a 29-real-number array and generates a single value output. The execution time of a single instance mono-core in an Intel i7-8550U @ 1.8GHz (up to 4.0GHz) is about 1 minute and 30 seconds. And after the change to multi-core (8 simulations in parallel), each execution time rises to 2 minutes.

When analysing clock time results, the CPU used and the \textit{pop size} parameter, which sets the limit of parallel samples executed, take an important role in the overall performance. In these cases, surrogates provide improvements: (1)~when the number of simulations is limited, but there is no limit in time, such as the case of competitions; or (2) more realistic scenarios when there is a hard limit in the number of fitness evaluations due to its individual execution time or the limitation in other resources, such as the number of licenses or GPU resources.  Besides, to make our analysis more realistic, we have also implemented a master/worker scheme for the parallel evaluation of the fitness function with several parallel threads. This configuration is also a common practice in simulation-based optimisation.    

When analysing the execution time for the three test configurations, DE complete execution, which is the baseline reference, takes an average of 11700s running with 8 threads of parallelism. Opposite, the DE + DT/C decreases the average to 9748s, running with the same 8 threads of parallelism, while getting better results (more discussion further). Nevertheless, some of the executions increase the overall execution time compared to DE. This increase relies on the situation when the surrogate model discards one or more samples, which can lead to some idle cores. One generation of the DE is formed by 15 samples, using 2 blocks of 8 threads (one CPU idle in each block). Although the surrogate leads to some fitness evaluation saving, one example generation of DE + DT/C with 12 samples (15 proposed by the DE and 3 discarded by the surrogate) would use the same 2 blocks of 8 threads, but this time 4 CPU will be idle, which do not turn into a significant reduction in execution time. In addition, this extra computational budget needs to be executed, which should improve the best quality sample found within the same budget and add some execution time. 


Further and alternatively, another execution time analysis includes extending the number of fitness executions of the DE until it reaches the values of DE + DT/C and comparing the difference in execution times to find a solution with the same quality. An equivalent analysis for benchmark tests was part of the comparison in Section~\ref{sec:computationaResourceUsage}. 

In addition, DE + XGB/R + Diver. execution time is significantly lower than DE + DT/C because it gets stuck into a situation in which the surrogate discards all new candidates (see section \ref{sec:noimprovementcondition}); therefore, the optimisation stops when there are 50 generations without improvement. It is, nonetheless, remarkable how DE + XGB/R reaches competitive results using less than one-third of the execution budget. 

\begin{figure}[!t]
    \centering
	\includegraphics[width=0.8\columnwidth, keepaspectratio=true]{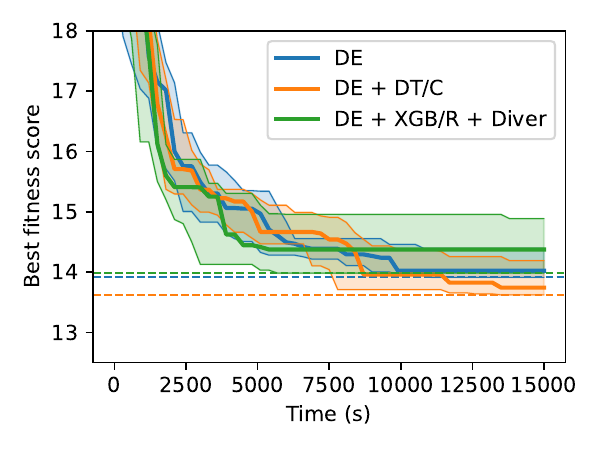}

	\caption{Confidence interval of the best quality score over the time. Figure shows the confidence interval within 0.25 and 0.75 percentiles plus the average in gross colour.}
	\label{fig:realWorldProgress}
\end{figure}

Besides, Figure \ref{fig:realWorldProgress} shows the progress of each configuration over time. Even though not statistically different, DE + DT/C shows a better behaviour for this problem than DE for all the percentiles at the end of the optimisation, consistent with obtained results during the benchmark analysis. Furthermore, looking at the lower percentiles, two phases can be identified. Firstly, DE + XGB/R + Diver. takes the lead until it starts discarding all the candidates. Secondly, DE + DT/C takes the lead as it keeps improving the solution, although the minimisation is slower than the DE + XGB/R + Neigh. As stated in section~\ref{sec:surrogateContributionEvolution}, this situation might suggest the existence of phases where one algorithm performs better than the rest depending on the stage of optimisation in which the algorithm and the population are.

\section{Conclusions}\label{sec:conclusiones}
This paper includes a multidimensional analysis of surrogate models under different configurations (e.g., different machine learning algorithms, different surrogate strategies, and two complementary problem formulations). We have proposed an alternative formulation for the traditional surrogate models that, instead of addressing the problem as a continuous variable prediction, considers how several optimisation algorithms use these models to redefine the problem as a binary class prediction model. Differential Evolution (DE) has been used as the baseline search strategy, which is both a method that allows this surrogate formulation (due to how new individuals are selected against their parents) and a well-performing technique in the repertoire of optimisation heuristics.

We have tackled the analysis of the surrogates following an online approach (alternating both the surrogate and the quality function during the optimisation) instead of offline (in which the quality function to optimise is replaced by the surrogate). The rationale behind selecting this approach is to conduct a more general study about how to control an optimisation, not limited to identifying the method that best fits a particular fitness landscape, which is a very problem-specific scenario.

The main contribution is the analysis of how adopting different machine learning models impacts the performance of a surrogate-based optimisation. The paper analyses both continuous and binary implementations corresponding, respectively, to surface and pairwise surrogate models. Furthermore, this study has also considered different strategies in applying surrogates, whether the surrogate filters candidate solutions only based on its predictions or considers other aspects such as quality distance or candidate diversity metrics. Several other authors, such as \citep{lim2009generalizing}, have mentioned that one of the ``\textit{greatest barrier to further progress is that with so many approximation techniques available in the literature, it is almost impossible to know which is the most relevant for modelling the problem landscape or generating reliable fitness predictions when one has only limited knowledge of its fitness space before the search starts}''. However, this paper included, in its analysis, a representative range of these general machine learning algorithms (regularisation-based regression, random forest, decision trees, neural networks, and boosting methods). Of course, there are many other techniques, but this is a representative sample of the existing methods, and this paper is one of the few studies in which several of them have been tested under different configurations to compare their effectiveness in the optimisation process. 

Even though, the experimental results have required the use of well-known benchmark problems instead of real-world problems. Authors, such as \citep{lim2009generalizing} stated that the ``\textit{observations obtained from the use of benchmark problems do not deviate significantly from those in the real-world problems we have experimented}''. Furthermore, Lurtis, the company for whom two of this paper's authors work, successfully uses these models for different real-world applications in science and engineering. 

As a result of the analysis, it is remarkable how some approaches effectively increase the performance, while other configurations negatively impact the overall optimisation process. The best performing configuration is the one using Decision Trees as a binary classifier (DT/C), followed by two configurations that used the traditional continuous variable prediction combined with a strategy to calculate solutions based on the diversity of the candidates (XGBoost boosting algorithm and Ridge regression, XGBoost/R+Diver. and Ridge/R + Diver). All the best surrogate configurations beat the baseline optimisation algorithm without surrogates, regardless of the machine learning algorithm or the surface/pairwise surrogate model. It is also interesting to observe how decision tree-based models pivot from worst to best performance when switching from a regression to a classification approach. Furthermore, the results have demonstrated that the diversity strategy is effective in boosting the performance of all the studied surrogate models with the exception of DT.  

The analysis of the results has also shown that the better performance of DT/C relies on its conservative behaviour, allowing it to exploit the optimisation algorithm for more generations while keeping constant the number of fitness evaluations required.This conclusion is consistent with the insights already highlighted by other authors, such as \citep{nguyen2014review}. The savings for achieving the same result are, in average, $>50\%$.  However, although DT/C has beaten the rest of the algorithms, it has also shown some scalability limitations due to the quadratic nature of the pairwise approach in long optimisation searches. A \textit{trail size} parameter has been introduced to achieve a trade-off between improvement in fitness and computational/memory costs.

As an additional contribution, this study has demonstrated that the standard quality measures used in machine learning (e.g., Mean Square Error or Accuracy) are not necessarily good indicators of the overall performance of an optimisation algorithm when using surrogates. This is in line with other authors, such as \citep{westermann2019surrogate}, who have suggested studies to unveil some insights into the behaviour of these methods in the optimisation process and the analysis of accuracy metrics. Specificity seems to be a good measure for algorithm performance in our study, as it somehow measures how conservative is an algorithm.

Besides, there are several aspects concerning the use of multiple and ideally heterogeneous models,  already suggested in \citep{jin2011surrogate}, that can ``\textit{considerably improve the search performance, compared to surrogate-assisted evolutionary algorithms}''. The results shown in this paper may indicate that there are several phases in which different surrogate models are contributing more to modelling the search process than others. The exploration of this hypothesis might suggest designing synergetic or self-adaptive approaches that might improve the results of any single surrogate model. 

Finally, the proposed alternative of formulating the surrogate model as a pairwise comparison rather than a continuous prediction opens the opportunity not only to use a different set of algorithms but also to stress some particular aspects of the behaviour of the surrogates that best complement the search mechanism. Among them, it is encouraged to study how the pairwise model can be connected to other search methods that do not involve pairwise search comparisons. Future studies will also cover the analysis of the surrogate models from the perspective of the dimensionality or complexity of the problem (as suggested in \citep{mallipeddi2015evolving}).  Also, this analysis opens new research lines in the dynamic configuration of the surrogate model acting according to the different optimisation phases.

\FloatBarrier

\bibliographystyle{apa}
\bibliography{surrogate}

\FloatBarrier

\appendix
\renewcommand{\thesection}{S \arabic{section}}

\renewcommand{\floatpagefraction}{0.99}%
\renewcommand{\topfraction}{0.99}
\renewcommand{\bottomfraction}{0.99}
\renewcommand{\dbltopfraction}{0.99}  

\section{Details of the machine learning algorithms}

All machine learning algorithms use the \textit{scikit-learn} python implementation\footnote{https://scikit-learn.org/} except for XGBoost/R that is based on its public Python implementation\footnote{https://xgboost.readthedocs.io/}. We have decided to keep all the parameters of the algorithms to their default values (no specific hyperparameter tunning) except for MLP/R, which has an initial learning rate of $0.01$ and it only performs $20$ iterations; and RF/R, which has only 40 estimators. In both cases, these values have been chosen to reduce the training time of the surrogates (see Table \ref{table:hyperparametersModels}).

\begin{table}[!htp]
    \centering
  \begin{tabular}{cp{4cm}}
    \hline
    \textbf{Model} & \textbf{Non-default configuration}\\
    \hline
        Ridge/R   & -    \\
        RF/R      & 'n\_estimators' = 40\\
        XGBoost/R & -                   \\
        DT/R      & -                   \\
        MLP/R     & 'learning\_rate\_init' = 0.01\hfill
                    'max\_iter' = 20\\
    \hline    
  \end{tabular}
  \caption{Non-default hyperparameters for models.}
  \label{table:hyperparametersModels}
\end{table}

\section{Details of the experimental scenario} \label{app:experimental-scenario}

For the experimental analysis of the different surrogate models, we have selected a well-known benchmark proposed in \citep{lozano2011editorial} for the SOCO'2011 special issue on the ``\textit{Scalability of evolutionary algorithms and other metaheuristics for large-scale continuous optimisation problems}''. This benchmark is made up of 19 continuous scalable functions with different characteristics. In order to adjust the experimentation to the inherent characteristics of the problems that are normally solved with the help of surrogate models, we have conducted the following changes to the usual experimental conditions of this benchmark.

\begin{itemize}
    \item Each configuration is run 15 times (instead of the usual 25) for each test function.

    \item The same run number for each of the strategies is guaranteed to start from the same initial population by using a common random seed. For a given strategy, different runs have randomly initialised populations.

    \item The maximum number of fitness evaluations has been fixed to 15D (i. e., 750), which is a much smaller number compared to the conditions of the original benchmark (5,000D). In similar studies \citep{Rehbach2018,Berveglieri2020}, the authors set a limit in the number of fitness function evaluations to 750 or 1250, respectively. This change is intended to mimic the kind of scenarios that are present in scientific and engineering problems, in which surrogate models are particularly useful. In these problems, the fitness function normally involves some kind of computer simulation (e.g., finite element or fluid dynamics solvers) that might take several minutes or even hours to run. Even if parallel evaluations are possible, it is not normally feasible to run such fitness functions a large number of times. Thus, in this study, even if benchmark functions and not simulations are considered, we have decided to scale down the budget of fitness evaluations to resemble this type of problems.
    
    \item Problem size has been fixed to $D=50$. This value ensures certain complexity while keeping computation time manageable for massive testing. Increasing $D$ value would mean increasing the maximum number of fitness evaluations as the budget is $f(D)$, which would make each optimisation process quite costly, but not necessarily more relevant. 
    
\end{itemize}   

The primary goal of this article is to analyse the potential of using surrogate models, based on different strategies, in order to save computation time while achieving a good-quality solution. It is not expected for any of the configurations to reach the optimal value in just 750 quality function evaluations, but to provide the best possible solution given the aforementioned budget. Additionally, this study aims to compare the pairwise surrogate modelling techniques (based on classifiers) with the surface (regression-based) equivalents. In order to face different models and strategies among them, the lowest quality value reached in the optimisation will be used as the comparison criterion. 

\section{Details of the reference algorithm} \label{app:reference-algorithm}

In particular, we have used the Differential Evolution Solver from the SciPy library as the reference implementation. This implementation includes a polishing phase that applies the L-BFGS-B algorithm after the DE that has been disabled in this experimentation (by setting the \texttt{intol} parameter to false). Additionally, a custom use of the \texttt{pop\_size} parameter has been implemented to be able to set a smaller population size value (15 in our case). This small population size has shown a good performance in the past for the same benchmark \citep{latorre2011mos} and is also consistent with the fact that we are considering an extremely low quality function evaluation budget. The whole set of hyperparameters is shown in Table~\ref{table:hyperparametersDE}.

\begin{table}[!htp]
    \centering
  \begin{tabular}{ccl}
    \hline
    \textbf{Parameter} & \textbf{Value} & \textbf{Description} \\
    \hline
    pop\_size          & 15             & Population size\\
    maxiter            & 50             & Max. num. generations \\
    strategy           & rand/1/exp     & Mutation strategy\\
    mutation           & 0.5            & Mutation factor\\
    recombination      & 0.5            & Recombination factor\\
    polish             & False          & Disable polishing phase \\
    \hline               
  \end{tabular}
    \caption{Hyperparameters for the DE algorithm.}
    \label{table:hyperparametersDE}
\end{table}

\section{Analysis of warm-up results.} \label{app:warm-up}

Tables~\ref{app:table:warmupTable} and \ref{app:table:warmupTablePairwise} show the results for both the \textit{surface} and the \textit{pairwise} surrogates, respectively, for each machine learning algorithm. The range of the warm-up values varies from 1 to 50 generations. In the extreme case of 50 generations, the surrogate model would only be active for the equivalent of one more generation of the DE algorithm. For comparison purposes, both tables also include the performance of the baseline DE algorithm with no surrogate. 

Table~\ref{app:table:warmupTable} presents the results for the \textit{surface surrogate} models. It is possible to observe that the performance of RF/R and DT/R seems to be worse (for all the warm-up values) than that of the baseline DE with no surrogate. These results could be considered enough to discard both approaches from further experimentation. However, for the sake of completeness, both algorithms have been kept for the following experiments with a \textit{warm-up phase} of 40 and 30 generations, respectively (the alternative of 50 generations has not been considered as there are almost no differences to not using a surrogate). On the other hand, XGBoost/R needs almost no warm-up to train a basic model, and the best performing configuration needs just 2 generations to start making good predictions. This allocates a large budget of quality function evaluations to the DE algorithm for the reminder of the search process. 

As shown in Table~\ref{app:table:warmupTablePairwise}, \textit{pairwise surrogate} models, in general, reduce the warm-up phase values compared to their \textit{surface surrogate} counterparts. In particular, MLP/C and DT/C
have dramatically reduced the warm-up duration for their respective best-performing configurations. Meanwhile, XGBoost/C has slightly increased the number of generations for the warm-up phase, but keeping it reasonably low. Finally, RF/C has also reduced the warm-up phase duration but, as opposed to the other ML algorithms, this value is still rather large. 

\begin{table*}
  \centering
  \begin{tabular}{@{}rrrrrrrrrr@{}}
    \multicolumn{2}{c}{RF/R} & \multicolumn{2}{c}{MLP/R}       &         \multicolumn{2}{c}{Ridge/R}     &         \multicolumn{2}{c}{DT/R} & \multicolumn{2}{c}{XGBoost/R}            \\ \hline
    WU          & Rank        & WU   & Rank & WU   & Rank & WU          & Rank        & WU   & Rank \\ \hline
    DE        & 3.79           & 30        & 5.84    & 10        & 7.00    & DE        & 7.58           & 2         & 5.58    \\
    50               & 4.84           & 40        & 6.11    & 7         & 7.21    & 30               & 8.00           & 1         & 5.74    \\
    40               & 5.89           & 20        & 6.37    & 5         & 7.53    & 50               & 9.47           & 3         & 5.74    \\
    30               & 7.21           & DE & 8.53    & 8         & 7.58    & 3                & 10.00          & 4         & 6.63    \\
    20               & 8.84           & 6         & 9.47    & 3         & 7.95    & 40               & 10.05          & 5         & 7.47    \\
    10               & 11.32          & 50        & 9.58    & 6         & 8.26    & 6                & 10.11          & 7         & 8.05    \\
    9                & 11.58          & 10        & 9.63    & 2         & 8.74    & 20               & 10.32          & 6         & 8.16    \\
    7                & 12.95          & 9         & 10.53   & 4         & 8.89    & 10               & 10.63          & 8         & 8.47    \\
    8                & 13.11          & 4         & 10.95   & 9         & 9.32    & 8                & 10.79          & 9         & 9.00    \\
    5                & 14.16          & 7         & 11.11   & 1         & 9.79    & 5                & 11.32          & 10        & 9.58    \\
    6                & 14.16          & 5         & 11.16   & 30        & 10.37   & 7                & 11.63          & 20        & 11.47   \\
    4                & 15.32          & 8         & 11.16   & 40        & 10.74   & 9                & 11.79          & 50        & 12.00   \\
    3                & 15.95          & 2         & 14.05   & 20        & 10.89   & 4                & 12.26          & 30        & 12.21   \\
    2                & 17.00          & 3         & 14.84   & 50        & 12.05   & 1                & 12.58          & 40        & 12.79   \\
    1                & 18.74          & 1         & 16.58   & DE & 12.74   & 2                & 13.16          & DE & 13.42  \\ \hline
  \end{tabular}
    \caption{Warm-up (WU) tuning results for the \textit{surface surrogate} strategy.}
    \label{app:table:warmupTable}
\end{table*}

\begin{table*}[!htp]
  \centering
  \begin{tabular}{@{}rrrrrrrrrr@{}}
    \multicolumn{2}{c}{RF/C}       & \multicolumn{2}{c}{MLP/C}                 & \multicolumn{2}{c}{Ridge/C}               & \multicolumn{2}{c}{DT/C}       & \multicolumn{2}{c}{XGBoost/C}             \\ \hline
    WU   & Rank & WU   & Rank & WU   & Rank & WU   & Rank & WU   & Rank \\ \hline
    20        & 6.05                        & 2         & 4.79                        & 2         & 6.00                        & 4         & 3.84                        & 6         & 6.00                        \\
    30        & 6.84                        & 4         & 5.58                        & 6         & 6.00                        & 3         & 3.95                        & 4         & 6.63                        \\
    8         & 7.05                        & 3         & 6.05                        & 3         & 6.42                        & 1         & 4.05                        & 8         & 6.63                        \\
    7         & 7.16                        & 6         & 6.05                        & 5         & 6.47                        & 2         & 4.16                        & 3         & 6.79                        \\
    DE & 7.29                        & 1         & 6.16                        & 10        & 6.68                        & 6         & 5.47                        & 9         & 7.37                        \\
    2         & 7.95                        & 7         & 7.05                        & 4         & 7.00                        & 5         & 6.11                        & 1         & 7.47                        \\
    50        & 8.08                        & 9         & 7.32                        & 1         & 7.16                        & 7         & 6.26                        & 7         & 7.84                        \\
    9         & 8.11                        & 8         & 7.47                        & 9         & 7.21                        & 9         & 6.53                        & 5         & 7.89                        \\
    1         & 8.63                        & 5         & 7.84                        & 7         & 7.26                        & 8         & 6.79                        & 10        & 7.95                        \\
    3         & 8.68                        & 10        & 8.26                        & 8         & 8.63                        & 10        & 8.63                        & 2         & 8.37                        \\
    6         & 8.68                        & DE & 10.45                       & 20        & 9.37                        & 20        & 10.53                       & 20        & 8.42                        \\
    10        & 8.68                        & 30        & 10.63                       & 30        & 9.95                        & 30        & 11.68                       & DE & 8.97                        \\
    4         & 8.74                        & 40        & 10.63                       & DE & 10.29                       & 40        & 13.11                       & 50        & 9.55                        \\
    5         & 8.79                        & 20        & 10.68                       & 40        & 10.63                       & DE & 14.18                       & 30        & 9.58                        \\
    40        & 9.26                        & 50        & 11.03                       & 50        & 10.92                       & 50        & 14.71                       & 40        & 10.53                       \\ \hline 
  \end{tabular}
  \caption{Warm-up (WU) tuning results for the \textit{pairwise surrogate} strategy.}
  \label{app:table:warmupTablePairwise}
\end{table*}

\section{Deeper analysis of the results}

This section provides a more detailed analysis of the results. \ref{app:sec:default_strategy} compares the models within its category (surface or pairwise) and all together, following the default strategy defined in Section~\ref{sec:baseline}. \ref{app:sec:strategies-analysis} compares for each model all proposed improvements (\textit{Probability, Quality and Diversity}). In addition, \ref{app:sec:best-sur} compares the models using its best strategy and \ref{app:sec:kriging-comparison} the winners with online and offline Kriging model.

\subsection{Default Strategy} \label{app:sec:default_strategy}
Table \ref{app:table:rankingSurfaceBaseline} shows the results of applying the default strategy ranked by performance. These results correspond to the surface surrogate model for all the machine learning algorithms under consideration.

\begin{table}[!htp]
  \centering
  \begin{tabular}{lc}
    \hline
    & Ranking \\ 
    \hline
    XGB/R & 2.37 \\ 
    Ridge/R & 2.63 \\ 
    MLP/R & 3.37 \\ 
    DT/R & 3.79 \\ 
    DE & 3.84 \\ 
    RF/R & 5.00 \\ 
    \hline
  \end{tabular}
  \caption{Average Ranking of Surface Baseline.}
  \label{app:table:rankingSurfaceBaseline}
\end{table}


Table \ref{app:table:validationSurfaceBaseline} displays the statistical comparison of the highest-ranked algorithm (XGBoost/R) with the rest of the surrogates using different machine learning algorithms. These results only show statistical differences with  RF/R, and only for the Friedman test. 

\begin{table}[!htp] 
  \centering 
  \begin{tabular}{ccccc} 
    \hline 
    XGBoost/R vs & Friedman$^*$ p-value & Wilcoxon$^*$ p-value \\  
    \hline 
    DE  & $6.07E-02$ & $1.82E-01$ \\  
    RF/R & $7.27E-05^{\surd}$ & $1.82E-01$ \\  
    Ridge/R & $6.65E-01$ & $8.87E-01$ \\  
    MLP/R & $1.99E-01$ & $1.82E-01$ \\  
    DT/R & $6.07E-02$ & $1.82E-01$ \\  
    \hline 
  \end{tabular} 
  \begin{tablenotes} 
    \item $\surd$ means that there are statistical differences with $\alpha=0.05$ 
    \item $^*$ means that the p-value has been corrected with the Holm procedure. 
  \end{tablenotes} 
  \caption{Statistical validation (XGBoost/R is the control algorithm).}  
  \label{app:table:validationSurfaceBaseline} 
\end{table} 

Secondly, Table~\ref{app:table:rankingPairwiseBaseline} presents the ranking of the pairwise surrogates counterparts, in which DT/C outperforms all the other approaches. The statistical significance analysis, shown in Table~\ref{app:table:validationPairwiseBaseline}, confirms this behaviour, compared to all the rest of the machine learning algorithms and for both statistical tests. 

\begin{table}[!htp]
  \centering
  \begin{tabular}{lc}
    \hline
    & Ranking \\ 
    \hline
    DT/C & 1.00 \\ 
    MLP/C & 2.63 \\ 
    Ridge/C & 3.05 \\ 
    XGBoost/C & 4.42 \\ 
    DE & 4.68 \\ 
    RF/C & 5.21 \\ 
    \hline
  \end{tabular}
    \caption{Average Ranking for Pairwise Baseline Approach.} 
    \label{app:table:rankingPairwiseBaseline}
\end{table}

\begin{table}[!htp] 
  \centering 
  \begin{tabular}{ccccc} 
    \hline 
    DT/C vs & Friedman$^*$ p-value & Wilcoxon$^*$ p-value \\  
    \hline 
    DE & $5.12E-09^{\surd}$ & $9.54E-06^{\surd}$ \\  
    MLP/C & $7.19E-03^{\surd}$ & $9.54E-06^{\surd}$ \\  
    RF/C & $2.00E-11^{\surd}$ & $9.54E-06^{\surd}$ \\  
    Ridge/C & $1.44E-03^{\surd}$ & $9.54E-06^{\surd}$ \\  
    XGBoost/C & $5.21E-08^{\surd}$ & $9.54E-06^{\surd}$ \\  
    \hline 
  \end{tabular} 
  \begin{tablenotes} 
    \item $\surd$ means that there are statistical differences with $\alpha=0.05$ 
    \item $^*$ means that the p-value has been corrected with the Holm procedure. 
  \end{tablenotes} 
    \caption{Statistical validation for Pairwise Baseline Approach (DT/C Pairwise is the control algorithm).}  
    \label{app:table:validationPairwiseBaseline} 
\end{table} 

Finally, and to conclude the default strategy analysis, Tables~\ref{app:table:rankingBaselineAll} and \ref{app:table:validationBaselineAll} compare the ranking and the statistical significance, respectively, for both the surface and the pairwise surrogate models. 

The results clearly show that the DT/C strategy not only outperforms all the other pairwise approaches but also all their surface counterparts. All the differences are statistically significant.

\begin{table}[!htp]
  \centering
  \begin{tabular}{lc}
    \hline
    & Ranking \\ 
    \hline
    DT/C & 1.16 \\ 
    XGBoost/R & 4.26 \\ 
    MLP/C & 4.37 \\ 
    Ridge/R & 4.68 \\ 
    Ridge/C & 5.21 \\ 
    XGBoost/C & 6.79 \\ 
    DE & 7.32 \\ 
    MLP/R & 7.32 \\ 
    DT/R & 7.42 \\ 
    RF/C & 8.32 \\ 
    RF/R & 9.16 \\ 
    \hline
  \end{tabular}
    \caption{Average Ranking for the surface and pairwise approaches without improvement strategies.} 
    \label{app:table:rankingBaselineAll}
\end{table}

\begin{table}[!htp] 
\centering 
\begin{tabular}{ccccc} 
  \hline 
DT/C vs & Friedman$^*$ p-value & Wilcoxon$^*$ p-value \\  
  \hline 
DE & $7.34E-08^{\surd}$ & $1.91E-05^{\surd}$ \\  
  DT/R & $4.69E-08^{\surd}$ & $1.91E-05^{\surd}$ \\  
  MLP/C & $5.70E-03^{\surd}$ & $1.91E-05^{\surd}$ \\  
  MLP/R & $7.34E-08^{\surd}$ & $1.91E-05^{\surd}$ \\  
  RF/C & $2.60E-10^{\surd}$ & $1.91E-05^{\surd}$ \\  
  RF/R & $1.05E-12^{\surd}$ & $1.91E-05^{\surd}$ \\  
  Ridge/C & $6.63E-04^{\surd}$ & $1.91E-05^{\surd}$ \\  
  Ridge/R & $3.15E-03^{\surd}$ & $1.67E-03^{\surd}$ \\  
  XGB/C & $8.31E-07^{\surd}$ & $1.91E-05^{\surd}$ \\  
  XGB/R & $5.70E-03^{\surd}$ & $1.91E-05^{\surd}$ \\  
   \hline 
\end{tabular} 
\begin{tablenotes} 
   \item $\surd$ means that there are statistical differences with $\alpha=0.05$ 
   \item $^*$ means that the p-value has been corrected with the Holm procedure. 
\end{tablenotes} 
\caption{Statistical validation for the surface and pairwise approaches without improvement strategies (DT/C is the control algorithm).}  
\label{app:table:validationBaselineAll} 
\end{table}

\subsection{Analysis of the Different Improved Strategies}\label{app:sec:strategies-analysis}

This second part of the experimentation covers the different alternative strategies presented in Section~\ref{sec:strategies}. This analysis takes each of these strategies (Default Strategy, Probability-based Strategy (\textit{Prob}), Quality Distance Strategy (\textit{Qual}), and Candidate Diversity Strategy (\textit{Diver})) and compares them independently and in combination. This type of analysis compares not only the individual contribution of each strategy but also the synergies resulting from their combination.

\subsubsection{Surface Surrogate Results}

Section \ref{app:sec:rfr} reports the results of the different strategies (and combinations) of the RF/R machine learning model. Analogously, sections \ref{app:sec:dtr}, \ref{app:sec:mlpr}, \ref{app:sec:ridger}, and \ref{app:sec:xgboostr} present the results for the DT/R, MLP/R, Ridge/R, and XGBoost/R models, respectively. To end \ref{app:sec:best-pair} shows the analysis among the best strategy for each algorithm.

\paragraph{Random Forest Results for Surface approach} \label{app:sec:rfr}

\begin{table}[H]
\label{app:table:ranking_rf_surface}
\centering
\begin{tabular}{lc}
  \hline
 & Ranking \\ 
  \hline
RF/R Diver. & 1.11 \\ 
  RF/R Diver. + Prob.  & 2.05 \\ 
  RF/R Diver. + Qual. & 4.18 \\ 
  DE & 5.13 \\ 
  RF/R Prob. & 5.63 \\ 
  RF/R All & 6.21 \\ 
  RF/R Qual. & 6.47 \\ 
  RF/R Prob. + Qual. & 6.95 \\ 
  RF/R  & 7.26 \\ 
   \hline
\end{tabular}
\caption{Average Ranking of RF/R for Surface approach.} 
\end{table}

\begin{table}[H] 
\centering 
\label{app:table:validation_rf_surface} 
\begin{tabular}{ccccc} 
  \hline 
RF/R Diver. vs & Friedman$^*$& Wilcoxon$^*$\\  
& p-value & p-value \\  

  \hline 
DE & $1.76E-05^{\surd}$ & $1.53E-05^{\surd}$ \\  
  RF/R & $3.36E-11^{\surd}$ & $2.61E-04^{\surd}$ \\  
  RF/R Prob. & $1.40E-06^{\surd}$ & $1.53E-05^{\surd}$ \\  
  RF/R Prob. + Qual. & $3.40E-10^{\surd}$ & $1.53E-05^{\surd}$ \\  
  RF/R Diver. + Prob.  & $2.86E-01$ & $1.53E-05^{\surd}$ \\  
  RF/R All & $4.57E-08^{\surd}$ & $1.53E-05^{\surd}$ \\  
  RF/R Qual. & $9.14E-09^{\surd}$ & $1.53E-05^{\surd}$ \\  
  RF/R Diver. + Qual. & $1.06E-03^{\surd}$ & $1.53E-05^{\surd}$ \\  
   \hline 
\end{tabular} 
\begin{tablenotes} 
   \item $\surd$ means that there are statistical differences with $\alpha=0.05$ 
   \item $^*$ means that the p-value has been corrected with the Holm procedure. 
\end{tablenotes} 
\caption{Statistical validation of RF/R for Surface approach (RF/R Diver. is the control algorithm).}  
\end{table} 

\paragraph{Decision Tree Results for Surface approach}\label{app:sec:dtr}
\begin{table}[H]
\label{app:table:ranking_dt_surface}
\centering
\begin{tabular}{lc}
  \hline
 & Ranking \\ 
  \hline
DT/R Diver. & 2.87 \\ 
  DT/R Diver. + Prob. & 3.47 \\ 
  DT/R Prob. + Qual. & 4.79 \\ 
  DT/R Diver. + Qual. & 5.11 \\ 
  DT/R  & 5.16 \\ 
  DT/R Qual. & 5.42 \\ 
  DT/R Prob. & 5.74 \\ 
  DT/R All & 6.00 \\ 
  DE & 6.45 \\ 
   \hline
\end{tabular}
\caption{Average Ranking of DT/R for Surface approach.} 
\end{table}

\begin{table}[H] 
\label{app:table:validation_dt_surface} 
\centering 
\begin{tabular}{ccccc} 
  \hline 
DT/R Diver. vs & Friedman$^*$ & Wilcoxon$^*$ \\  
& p-value & p-value \\  
  \hline 
DE & $4.50E-04^{\surd}$ & $4.68E-03^{\surd}$ \\  
  DT/R  & $3.99E-02^{\surd}$ & $3.38E-01$ \\  
  DT/R Prob. & $7.47E-03^{\surd}$ & $3.38E-01$ \\  
  DT/R Prob. + Qual. & $6.12E-02$ & $1.51E-01$ \\  
  DT/R Diver. + Prob. & $4.96E-01$ & $3.38E-01$ \\  
  DT/R All & $2.97E-03^{\surd}$ & $2.50E-02^{\surd}$ \\  
  DT/R Qual. & $2.03E-02^{\surd}$ & $2.89E-01$ \\  
  DT/R Diver. + Qual. & $3.99E-02^{\surd}$ & $1.08E-01$ \\  
   \hline 
\end{tabular} 
\begin{tablenotes} 
   \item $\surd$ means that there are statistical differences with $\alpha=0.05$ 
   \item $^*$ means that the p-value has been corrected with the Holm procedure. 
\end{tablenotes} 
\caption{Statistical validation of DT/R for Surface approach (DT/R Diver. is the control algorithm).}  
\end{table}

\paragraph{Multi-Layer Perceptron Results for Surface approach}\label{app:sec:mlpr}
\begin{table}[H]
\label{app:table:ranking_mlp_surface}
\centering
\begin{tabular}{lc}
  \hline
 & Ranking \\ 
  \hline
MLP/R Diver. & 1.05 \\ 
  MLP/R Diver. + Prob. & 2.11 \\ 
  MLP/R  & 4.32 \\ 
  MLP/R Prob. & 4.84 \\ 
  DE & 5.18 \\ 
  MLP/R Diver. + Qual. & 5.18 \\ 
  MLP/R Prob. + Qual. & 6.74 \\ 
  MLP/R All & 7.47 \\ 
  MLP/R Qual. & 8.11 \\ 
   \hline
\end{tabular}
\caption{Average Ranking of MLP/R for Surface approach.} 
\end{table}

\begin{table}[H]
\label{app:table:validation_mlp_surface} 
\centering 
\begin{tabular}{ccccc} 
  \hline 
MLP/R Diver. vs & Friedman$^*$ & Wilcoxon$^*$ \\  
& p-value & p-value \\  

  \hline 
DE & $1.66E-05^{\surd}$ & $1.53E-05^{\surd}$ \\  
  MLP/R  & $4.80E-04^{\surd}$ & $1.53E-05^{\surd}$ \\  
  MLP/R Prob. & $6.00E-05^{\surd}$ & $1.53E-05^{\surd}$ \\  
  MLP/R Prob. + Qual. & $9.48E-10^{\surd}$ & $1.53E-05^{\surd}$ \\  
  MLP/R Diver. + Prob. & $2.36E-01$ & $5.86E-04^{\surd}$ \\  
  MLP/R All & $3.46E-12^{\surd}$ & $1.53E-05^{\surd}$ \\  
  MLP/R Qual.& $1.60E-14^{\surd}$ & $1.53E-05^{\surd}$ \\  
  MLP/R Diver. + Qual. & $1.66E-05^{\surd}$ & $1.53E-05^{\surd}$ \\  
   \hline 
\end{tabular} 
\begin{tablenotes} 
   \item $\surd$ means that there are statistical differences with $\alpha=0.05$ 
   \item $^*$ means that the p-value has been corrected with the Holm procedure. 
\end{tablenotes} 
\caption{Statistical validation of MLP/R for Surface approach (MLP/R Diver. is the control algorithm).}  
\end{table} 

\paragraph{Ridge Results for Surface approach}\label{app:sec:ridger}
\begin{table}[H]
\label{app:table:ranking_ridge_surface}
\centering
\begin{tabular}{lc}
  \hline
 & Ranking \\ 
  \hline
Ridge/R Diver. & 1.53 \\ 
  Ridge/R Diver. + Prob.  & 2.79 \\ 
  Ridge/R  & 3.53 \\ 
  Ridge/R Prob. & 4.11 \\ 
  Ridge/R Diver. + Qual. & 5.71 \\ 
  DE & 5.82 \\ 
  Ridge/R Prob. + Qual. & 6.84 \\ 
  Ridge/R All & 7.11 \\ 
  Ridge/R Qual. & 7.58 \\ 
   \hline
\end{tabular}
\caption{Average Ranking of Ridge/R for Surface approach.} 
\end{table}

\begin{table}[H] 
\label{app:table:validation_ridge_surface} 
\centering 
\begin{tabular}{ccccc} 
  \hline 
Ridge/R Diver. vs & Friedman$^*$ & Wilcoxon$^*$ \\ 
& p-value & p-value \\  

  \hline 
DE & $6.91E-06^{\surd}$ & $1.53E-05^{\surd}$ \\  
  Ridge/R  & $4.88E-02^{\surd}$ & $5.08E-01$ \\  
  Ridge/R Prob. & $1.11E-02^{\surd}$ & $1.53E-05^{\surd}$ \\  
  Ridge/R Prob. + Qual. & $1.32E-08^{\surd}$ & $1.53E-05^{\surd}$ \\  
  Ridge/R Diver. + Prob.  & $1.55E-01$ & $1.17E-03^{\surd}$ \\  
  Ridge/R All & $2.39E-09^{\surd}$ & $1.53E-05^{\surd}$ \\  
  Ridge/R Qual. & $7.70E-11^{\surd}$ & $1.53E-05^{\surd}$ \\  
  Ridge/R Diver. + Qual. & $9.95E-06^{\surd}$ & $1.53E-05^{\surd}$ \\  
   \hline 
\end{tabular} 
\begin{tablenotes} 
   \item $\surd$ means that there are statistical differences with $\alpha=0.05$ 
   \item $^*$ means that the p-value has been corrected with the Holm procedure. 
\end{tablenotes}
\caption{Statistical validation of Ridge/R for Surface approach (Ridge/R Diver. is the control algorithm).}  
\end{table}

\paragraph{XGBoost Results for Surface approach}\label{app:sec:xgboostr}
\begin{table}[H]
\label{app:table:ranking_xgb_surface}
\centering
\begin{tabular}{lc}
  \hline
 & Ranking \\ 
  \hline
XGBoost/R Diver. & 1.47 \\ 
  XGBoost/R Diver. + Prob. & 2.58 \\ 
  XGBoost/R  & 3.11 \\ 
  XGBoost/R Prob. & 4.42 \\ 
  XGBoost/R Diver. + Qual. & 5.55 \\ 
  DE & 5.82 \\ 
  XGBoost/R Prob. + Qual. & 6.58 \\ 
  XGBoost/R All & 7.58 \\ 
  XGBoost/R Qual.& 7.89 \\ 
   \hline
\end{tabular}
\caption{Average Ranking of XGBoost/R for Surface approach.} 
\end{table}

\begin{table}[H] 
\label{app:table:validation_xgb_surface} 
\centering 
\begin{tabular}{ccccc} 
  \hline 
XGBoost/R Diver. vs & Friedman$^*$ & Wilcoxon$^*$ \\  
& p-value & p-value \\  

  \hline 
DE & $5.12E-06^{\surd}$ & $1.53E-05^{\surd}$ \\  
  XGBoost/R & $1.33E-01$ & $1.66E-01$ \\  
  XGBoost/R Prob. & $2.73E-03^{\surd}$ & $1.53E-05^{\surd}$ \\  
  XGBoost/R Prob. + Qual. & $5.49E-08^{\surd}$ & $1.53E-05^{\surd}$ \\  
  XGBoost/R Diver. + Prob. & $2.14E-01$ & $1.69E-03^{\surd}$ \\  
  XGBoost/R All & $4.45E-11^{\surd}$ & $1.53E-05^{\surd}$ \\  
  XGBoost/R Qual.& $3.96E-12^{\surd}$ & $1.53E-05^{\surd}$ \\  
  XGBoost/R Diver. + Qual. & $1.77E-05^{\surd}$ & $1.53E-05^{\surd}$ \\  
   \hline 
\end{tabular} 
\begin{tablenotes} 
   \item $\surd$ means that there are statistical differences with $\alpha=0.05$ 
   \item $^*$ means that the p-value has been corrected with the Holm procedure. 
\end{tablenotes} 
\caption{Statistical validation of XGBoost/R for Surface approach (XGBoost/R Diver. is the control algorithm).}  
\end{table}

\paragraph{Best combination Results for Surface approach}\label{app:sec:best-sur}

Among all the combinations tested, the best performing ones are those using the Candidate Diversity Strategy (\textit{Diver}). A summary of these results is presented in Table~\ref{app:table:rankingAllImpSurface}, which shows the comparative ranking of the best strategy for each machine learning algorithm. Besides, Table~\ref{app:table:validationAllImpSurface} displays the corresponding statistical analysis, in which the  XGBoost/R+Diver. combination is used as the reference algorithm, as it obtained the best overall results.

\begin{table}[!htp]
\centering
\begin{tabular}{lc}
  \hline
 & Ranking \\ 
  \hline
XGBoost/R + Diver. & 1.16 \\ 
  Ridge/R + Diver. & 1.89 \\ 
  MLP/R + Diver. & 3.05 \\ 
  RF/R + Diver. & 3.97 \\ 
  DT/R + Diver. & 4.97 \\ 
  DE & 5.95 \\ 
   \hline
\end{tabular}
\caption{Average Ranking for all the machine learning models with their respective best strategy for the surface approach.} 
\label{app:table:rankingAllImpSurface}
\end{table}

The statistical analysis, in this case, reveals significant differences between XGBoost/R+Diver. and the best performing strategies for the other machine learning algorithms. The only exception is Ridge/R+Diver, which is a rather simple machine learning model, but not performing much worse than the reference algorithm (and only for the Friedman test).

\begin{table}[!htp] 
\centering 
\begin{tabular}{ccccc} 
  \hline 
XGBoost/R & Friedman$^*$ & Wilcoxon$^*$  \\  
+ Diver. vs & p-value & p-value \\  
  \hline 
DE & $1.55E-14^{\surd}$ & $9.54E-06^{\surd}$ \\  
  RF/R + Diver. & $1.05E-05^{\surd}$ & $9.54E-06^{\surd}$ \\  
  Ridge/R + Diver. & $2.25E-01$ & $1.01E-03^{\surd}$ \\  
  MLP/R + Diver. & $3.60E-03^{\surd}$ & $1.14E-05^{\surd}$ \\  
  DTR/R + Diver. & $1.30E-09^{\surd}$ & $9.54E-06^{\surd}$ \\  
   \hline 
\end{tabular} 
\begin{tablenotes} 
   \item $\surd$ means that there are statistical differences with $\alpha=0.05$ 
   \item $^*$ means that the p-value has been corrected with the Holm procedure. 
\end{tablenotes} 
\caption{Statistical validation for all the models with their best strategy for the surface approach (XGBoost/R + Diver. is the control algorithm).}  
\label{app:table:validationAllImpSurface} 
\end{table}

\subsubsection{Pairwise Surrogate Results}
In a similar way as shown in the case of the Surface Surrogate Models, sections \ref{app:sec:rfc}, \ref{app:sec:dtc}, \ref{app:sec:mlpc}, \ref{app:sec:ridgec}, and \ref{app:sec:xgboostc} present the results for the pairwise models based on the RF/C, DT/C, MLP/C, Ridge/C, and XGBoost/C machine learning models, respectively. To end \ref{app:sec:best-pair} shows the analysis among the best strategy for each algorithm.

\paragraph{Random Forest Results for Pairwise approach}\label{app:sec:rfc}
\begin{table}[H]
\label{app:table:ranking_rf_pairwise}
\centering
\begin{tabular}{lc}
  \hline
 & Ranking \\ 
  \hline
RF/C Diver. & 1.42 \\ 
  RF/C Diver. + Qual. & 3.16 \\ 
  RF/C Diver. + Prob. & 3.37 \\ 
  RF/C All & 4.11 \\ 
  RF/C Qual.& 6.05 \\ 
  RF/C Prob. & 6.16 \\ 
  RF/C Prob. + Qual. & 6.63 \\ 
  RF/C & 6.84 \\ 
  DE & 7.26 \\ 
   \hline
\end{tabular}
\caption{Average Ranking of RF/C for Pairwise approach.} 
\end{table}

\begin{table}[H] 
\label{app:table:validation_rf_pairwise} 
\centering 
\begin{tabular}{ccccc} 
  \hline 
RF/C Diver. vs & Friedman$^*$ & Wilcoxon$^*$ \\ 
& p-value & p-value \\  

  \hline 
  DE & $3.89E-10^{\surd}$ & $1.53E-05^{\surd}$ \\  
  RF/C & $7.37E-09^{\surd}$ & $1.53E-05^{\surd}$ \\  
  RF/C Prob. & $4.88E-07^{\surd}$ & $4.12E-03^{\surd}$ \\  
  RF/C Prob. + Qual. & $2.71E-08^{\surd}$ & $1.29E-03^{\surd}$ \\  
  RF/C Diver. + Prob. & $5.68E-02$ & $1.69E-03^{\surd}$ \\  
  RF/C All & $7.56E-03^{\surd}$ & $1.53E-05^{\surd}$ \\  
  RF/C Qual.& $7.45E-07^{\surd}$ & $1.29E-03^{\surd}$ \\  
  RF/C Diver. + Qual. & $5.68E-02$ & $5.25E-04^{\surd}$ \\  
   \hline 
\end{tabular} 
\begin{tablenotes} 
   \item $\surd$ means that there are statistical differences with $\alpha=0.05$ 
   \item $^*$ means that the p-value has been corrected with the Holm procedure. 
\end{tablenotes} 
\caption{Statistical validation of RF/C for Pairwise approach(RF/C Diver. is the control algorithm).}  
\end{table}

\paragraph{Decision Tree Results for Pairwise approach}\label{app:sec:dtc}
\begin{table}[H]
\label{app:table:ranking_dt_pairwise}
\centering
\begin{tabular}{lc}
  \hline
 & Ranking \\ 
  \hline
DT/C  & 1.16 \\ 
  DT/C Diver. & 2.16 \\ 
  DT/C Diver. + Prob. & 3.74 \\ 
  DT/C Prob.& 4.16 \\ 
  DT/C Diver. + Qual. & 5.16 \\ 
  DT/C Qual. & 5.68 \\ 
  DT/C All & 6.26 \\ 
  DT/C Prob. + Qual. & 7.89 \\ 
  DE & 8.79 \\ 
   \hline
\end{tabular}
\caption{Average Ranking of DT/C for Pairwise approach.} 
\end{table}

\begin{table}[H] 
\label{app:table:validation_dt_pairwise} 
\centering 
\begin{tabular}{ccccc} 
  \hline 
DT/C  vs & Friedman$^*$ & Wilcoxon$^*$ \\
& p-value & p-value \\  

  \hline 
DE & $0.00E+00^{\surd}$ & $1.53E-05^{\surd}$ \\  
  DT/C Prob.& $2.20E-03^{\surd}$ & $1.53E-05^{\surd}$ \\  
  DT/C Prob. + Qual. & $2.38E-13^{\surd}$ & $1.53E-05^{\surd}$ \\  
  DT/C Diver. + Prob. & $7.40E-03^{\surd}$ & $1.53E-05^{\surd}$ \\  
  DT/C All & $5.49E-08^{\surd}$ & $1.53E-05^{\surd}$ \\  
  DT/C Diver. & $2.60E-01$ & $2.29E-03^{\surd}$ \\  
  DT/C Qual. & $1.75E-06^{\surd}$ & $1.53E-05^{\surd}$ \\  
  DT/C Diver. + Qual. & $2.69E-05^{\surd}$ & $1.53E-05^{\surd}$ \\  
   \hline 
\end{tabular} 
\begin{tablenotes} 
   \item $\surd$ means that there are statistical differences with $\alpha=0.05$ 
   \item $^*$ means that the p-value has been corrected with the Holm procedure. 
\end{tablenotes} 
\caption{Statistical validation of DT/C for Pairwise approach(DT/C is the control algorithm).}  
\end{table}

\paragraph{Multi-Layer Perceptron Results for Pairwise approach}\label{app:sec:mlpc}
\begin{table}[H]
\label{app:table:ranking_mlp_pairwise}
\centering
\begin{tabular}{lc}
  \hline
 & Ranking \\ 
  \hline
MLP/C Diver. & 2.79 \\ 
  MLP/C  & 3.95 \\ 
  MLP/C Prob. & 4.32 \\ 
  MLP/C Diver. + Qual. & 4.32 \\ 
  MLP/C Diver. + Prob. & 4.37 \\ 
  MLP/C Qual. & 4.68 \\ 
  MLP/C All & 5.95 \\ 
  MLP/C Prob. + Qual. & 6.63 \\ 
  DE & 8.00 \\ 
   \hline
\end{tabular}
\caption{Average Ranking of MLP/C for Pairwise approach.} 
\end{table}

\begin{table}[H] 
\label{app:table:validation_mlp_pairwise} 
\centering 
\begin{tabular}{ccccc} 
  \hline 
MLP/C Diver. vs & Friedman$^*$ & Wilcoxon$^*$ \\
& p-value & p-value \\  

  \hline 
DE & $3.61E-08^{\surd}$ & $1.53E-05^{\surd}$ \\  
  MLP/C  & $3.02E-01$ & $8.29E-01$ \\  
  MLP/C Prob. & $3.02E-01$ & $8.29E-01$ \\  
  MLP/C Prob. + Qual. & $1.07E-04^{\surd}$ & $2.06E-02^{\surd}$ \\  
  MLP/C Diver. + Prob. & $3.02E-01$ & $1.93E-03^{\surd}$ \\  
  MLP/C All & $2.28E-03^{\surd}$ & $1.53E-05^{\surd}$ \\  
  MLP/C Qual. & $1.65E-01$ & $7.73E-01$ \\  
  MLP/C Diver. + Qual. & $3.02E-01$ & $1.75E-01$ \\  
   \hline 
\end{tabular} 
\begin{tablenotes} 
   \item $\surd$ means that there are statistical differences with $\alpha=0.05$ 
   \item $^*$ means that the p-value has been corrected with the Holm procedure. 
\end{tablenotes} 
\caption{Statistical validation of MLP/C for Pairwise approach(MLP/C Diver. is the control algorithm).}  
\end{table}

\paragraph{Ridge Results for Pairwise approach}\label{app:sec:ridgec}
\begin{table}[H]
\label{app:table:ranking_ridge_pairwise}
\centering
\begin{tabular}{lc}
  \hline
 & Ranking \\ 
  \hline
Ridge/C Diver. & 1.89 \\ 
  Ridge/C Diver. + Prob. & 3.05 \\ 
  Ridge/C  & 3.16 \\ 
  Ridge/C Prob. & 3.42 \\ 
  Ridge/C All & 5.45 \\ 
  Ridge/C Diver. + Qual. & 5.87 \\ 
  DE & 7.00 \\ 
  Ridge/C Prob. + Qual. & 7.21 \\ 
  Ridge/C Qual.& 7.95 \\ 
   \hline
\end{tabular}
\caption{Average Ranking of Ridge/C for Pairwise approach.} 
\end{table}

\begin{table}[H] 
\centering 
\label{app:table:validation_ridge_pairwise} 
\begin{tabular}{ccccc} 
  \hline 
Ridge/C Diver. vs & Friedman$^*$ & Wilcoxon$^*$ \\
& p-value & p-value \\  

  \hline 
DE & $5.49E-08^{\surd}$ & $1.53E-05^{\surd}$ \\  
  Ridge/C  & $3.10E-01$ & $6.75E-01$ \\  
  Ridge/C Prob. & $2.57E-01$ & $3.87E-02^{\surd}$ \\  
  Ridge/C Prob. + Qual. & $1.54E-08^{\surd}$ & $1.53E-05^{\surd}$ \\  
  Ridge/C Diver. + Prob. & $3.10E-01$ & $3.87E-02^{\surd}$ \\  
  Ridge/C All & $2.55E-04^{\surd}$ & $1.53E-05^{\surd}$ \\  
  Ridge/C Qual.& $7.70E-11^{\surd}$ & $1.53E-05^{\surd}$ \\  
  Ridge/C Diver. + Qual. & $3.87E-05^{\surd}$ & $1.53E-05^{\surd}$ \\  
   \hline 
\end{tabular} 
\begin{tablenotes} 
   \item $\surd$ means that there are statistical differences with $\alpha=0.05$ 
   \item $^*$ means that the p-value has been corrected with the Holm procedure. 
\end{tablenotes} 
\caption{Statistical validation of Ridge/C for Pairwise approach(Ridge/C Diver. is the control algorithm).}  
\end{table}

\paragraph{XGBoost Results for Pairwise approach}\label{app:sec:xgboostc}
\begin{table}[H]
\label{app:table:ranking_xgb_pairwise}
\centering
\begin{tabular}{lc}
  \hline
 & Ranking \\ 
  \hline
XGBoost/C Diver. & 2.00 \\ 
  XGBoost/C Diver. + Prob.  & 3.05 \\ 
  XGBoost/C Diver. + Qual. & 3.63 \\ 
  XGBoost/C & 4.32 \\ 
  XGBoost/C All & 4.89 \\ 
  XGBoost/C Prob. & 5.84 \\ 
  XGBoost/C Qual. & 6.32 \\ 
  XGBoost/C Prob. + Qual. & 7.42 \\ 
  DE & 7.53 \\ 
   \hline
\end{tabular}
\caption{Average Ranking of XGBoost/C for Pairwise approach.} 
\end{table}

\begin{table}[H] 
\label{app:table:validation_xgb_pairwise} 
\centering 
\begin{tabular}{ccccc} 
  \hline 
XGBoost/C Diver. vs & Friedman$^*$ & Wilcoxon$^*$\\  
& p-value & p-value \\  

  \hline 
DE & $3.99E-09^{\surd}$ & $1.53E-05^{\surd}$ \\  
  XGBoost/C & $2.75E-02^{\surd}$ & $2.97E-01$ \\  
  XGBoost/C Prob. & $7.66E-05^{\surd}$ & $6.68E-05^{\surd}$ \\  
  XGBoost/C Prob. + Qual. & $7.37E-09^{\surd}$ & $1.53E-05^{\surd}$ \\  
  XGBoost/C Diver. + Prob.  & $2.36E-01$ & $1.70E-01$ \\  
  XGBoost/C All & $4.49E-03^{\surd}$ & $1.53E-05^{\surd}$ \\  
  XGBoost/C Qual. & $7.14E-06^{\surd}$ & $6.71E-04^{\surd}$ \\  
  XGBoost/C Diver. + Qual. & $1.33E-01$ & $1.82E-01$ \\  
   \hline 
\end{tabular} 
\begin{tablenotes} 
   \item $\surd$ means that there are statistical differences with $\alpha=0.05$ 
   \item $^*$ means that the p-value has been corrected with the Holm procedure. 
\end{tablenotes} 
\caption{Statistical validation of XGBoost/C for Pairwise approach (XGBoost/C Diver. is the control algorithm).}  
\end{table} 

\paragraph{Best combination Results for Pairwise approach}\label{app:sec:best-pair}

A quick analysis of these results reveals that the Candidate Diversity Strategy (\textit{Diver}) is, again, the best performing one for most of the machine learning algorithms, as it was for the Surface Surrogate Models. The only exception here is the DT/C model (without any strategy) which, surprisingly, obtained the best ranking (see Table~\ref{app:table:rankingAllImpPairwise}).

\begin{table}[!htp]
\centering
\begin{tabular}{lc}
  \hline
 & Ranking \\ 
  \hline
DT/C & 1.00 \\ 
  Ridge/C + Diver. & 2.79 \\ 
  MLP/C + Diver. & 2.95 \\ 
  XGBoost/C + Diver. & 3.42 \\ 
  RF/C + Diver. & 4.84 \\ 
  DE & 6.00 \\ 
   \hline
\end{tabular}
\caption{Average Ranking for all the models with their best strategy for the Pairwise approach.} 
\label{app:table:rankingAllImpPairwise}
\end{table}

Table~\ref{app:table:validationAllImpPairwise} also reports significant differences for the statistical validation between DT/C, the reference algorithm (best ranking) and all the other combinations of machine learning algorithms and strategies.

\begin{table}[!htp] 
\centering 
\begin{tabular}{ccccc} 
  \hline 
DT/C vs & Friedman$^*$ & Wilcoxon$^*$ \\  
 & p-value &  p-value \\ 
  \hline 
DE & $1.11E-15^{\surd}$ & $9.54E-06^{\surd}$ \\  
  MLP/C + Diver. & $2.67E-03^{\surd}$ & $9.54E-06^{\surd}$ \\  
  RF/C + Diver. & $9.81E-10^{\surd}$ & $9.54E-06^{\surd}$ \\  
  Ridge/C + Diver. & $3.20E-03^{\surd}$ & $9.54E-06^{\surd}$ \\  
  XGBoost/C + Diver. & $1.99E-04^{\surd}$ & $9.54E-06^{\surd}$ \\  
   \hline 
\end{tabular} 
\begin{tablenotes} 
   \item $\surd$ means that there are statistical differences with $\alpha=0.05$ 
   \item $^*$ means that the p-value has been corrected with the Holm procedure. 
\end{tablenotes} 
\caption{Statistical validation for all the models with their best strategy for the Pairwise approach (DT/C is the control algorithm).}  
\label{app:table:validationAllImpPairwise} 

\end{table}

\subsubsection{Pairwise vs Surface Surrogate Results}\label{app:sec:final-comparison}

This final analysis compares the best combination of machine learning model and strategy (\textit{Diver} in most of the cases) for both the surface and the pairwise models. The ranking of this overall comparison, shown in Table \ref{app:table:rankingAllImpBoth}, indicates that the best surrogate configuration is DT/C. By just observing the ranking values, it seems clear that there is a second group of algorithms, including XGBoost/R+Diver. and Ridge/R+Diver., with a similar performance but worse than DT/C. The third group of algorithms includes Ridge/C+Diver, MLP/C+Diver, MLP/R+Diver, and XGBoost/C+Diver, all of them with very similar rankings. At the bottom of the ranking we can find RF/C+Diver, RF/R+Diver. and DT/R+Diver. (being the last one slightly worse than the other two). Finally, the baseline algorithm, the DE with no surrogate model, is the worst alternative.

\begin{table}[!htp]
\centering
\begin{tabular}{lc}
  \hline
 & Ranking \\ 
  \hline
  DT/C & 1.16 \\ 
  XGBoost/R + Diver. & 2.21 \\ 
  Ridge/R + Diver. & 2.89 \\ 
  Ridge/C + Diver. & 4.95 \\ 
  MLP/C + Diver. & 5.32 \\ 
  MLP/R + Diver. & 5.89 \\ 
  XGBoost/C + Diver. & 5.95 \\ 
  RF/C + Diver. & 8.16 \\ 
  RF/R + Diver. & 8.61 \\ 
  DT/R + Diver. & 9.92 \\ 
  DE & 10.95 \\ 
   \hline
\end{tabular}
\caption{Average Ranking.} 
\label{app:table:rankingAllImpBoth}
\end{table}

Table~\ref{app:table:validationAllImpBoth} reveals significant differences in the statistical comparison between the reference algorithm (DT/C) and all the other approaches for both statistical tests, with the only exceptions of
Ridge/R+Diver. and XGBoost/R+Diver. and the Friedman test.

\begin{table}[!htp] 
\centering 
\begin{tabular}{ccccc} 
  \hline 
 & Friedman$^*$ & Wilcoxon$^*$ \\  
DT/C vs & p-value & p-value \\  
  \hline 
DE & $0.00E+00^{\surd}$ & $1.91E-05^{\surd}$ \\  
  MLP/C + Diver. & $4.46E-04^{\surd}$ & $1.91E-05^{\surd}$ \\  
  RF/C + Diver. & $5.43E-10^{\surd}$ & $1.91E-05^{\surd}$ \\  
  Ridge/C + Diver. & $1.29E-03^{\surd}$ & $1.91E-05^{\surd}$ \\  
  XGBoost/C + Diver. & $5.13E-05^{\surd}$ & $1.91E-05^{\surd}$ \\  
  RF/R + Diver. & $3.59E-11^{\surd}$ & $1.91E-05^{\surd}$ \\  
  Ridge/R + Diver. & $2.13E-01$ & $2.10E-04^{\surd}$ \\  
  MLP/R + Diver. & $5.36E-05^{\surd}$ & $1.91E-05^{\surd}$ \\  
  DT/R + Diver. & $4.00E-15^{\surd}$ & $1.91E-05^{\surd}$ \\  
  XGBoost/R + Diver. & $3.28E-01$ & $2.10E-04^{\surd}$ \\  
   \hline 
\end{tabular} 
\begin{tablenotes} 
   \item $\surd$ means that there are statistical differences with $\alpha=0.05$ 
   \item $^*$ means that the p-value has been corrected with the Holm procedure. 
\end{tablenotes} 
\caption{Statistical validation (DT/C is the control algorithm).}  \label{app:table:validationAllImpBoth} 
\end{table}

\subsubsection{Best approaches vs Kriging results}\label{app:sec:kriging-comparison}

Finally, the best surrogates found in the final comparison are faced against one of the most used surrogate, Kriging model. For such purpose, two alternatives are considered: online and offline surrogate. Table \ref{app:table:rankingKriging} shows the obtained ranking, where the online approach gets a better result than Ridge/R + Diver. Meanwhile, Table \ref{app:table:validationKriging} depicts the statistical comparison versus the reference algorithm (DT/C). Results showed that DT/C performed significantly better than all the other approaches, including Kriging model.

\begin{table}[!htp]
\centering
\begin{tabular}{lc}
  \hline
 & Ranking \\ 
  \hline
DT/C & 1.63 \\ 
  XGBoost/R + Diver. & 2.89 \\ 
  Krigging (Online) & 3.11 \\ 
  Ridge/R + Diver. & 3.89 \\ 
  Kriging (Offline) & 3.89 \\ 
  Benchmark & 5.58 \\ 
   \hline
\end{tabular}
\caption{Average ranking for the found solutions vs Kriging models} 
\label{app:table:rankingKriging}
\end{table}

\begin{table}[!htp] 
\centering 
\begin{tabular}{ccccc} 
  \hline 
  \hline 
 & Friedman$^*$ & Wilcoxon$^*$ \\  
DT/C vs & p-value & p-value \\   
  \hline 
Benchmark  & $3.93E-10^{\surd}$ & $9.54E-06^{\surd}$ \\  
  Ridge/R + Diver. & $7.70E-04^{\surd}$ & $4.20E-04^{\surd}$ \\  
  XGBoost/R + Diver.  & $3.74E-02^{\surd}$ & $4.20E-04^{\surd}$ \\  
  Krigging (Online)  & $3.04E-02^{\surd}$ & $1.69E-03^{\surd}$ \\  
  Kriging (Offline)  & $7.70E-04^{\surd}$ & $2.23E-02^{\surd}$ \\  
   \hline 
\end{tabular} 
\begin{tablenotes} 
   \item $\surd$ means that there are statistical differences with significance level $\alpha=0.05$ 
   \item $^*$ means that the p-value has been corrected with the Holm procedure. 
\end{tablenotes} 
\caption{Statistical validation (DT/C is the control algorithm)}  
\label{app:table:validationKriging} 
\end{table}

\onecolumn
\FloatBarrier

\section{Complete Accuracy Analysis}\label{app:sec:all-accuracy}

\begin{figure*}[!b]
    \centering
  \includegraphics[width=\textwidth, height = 0.88\textheight,keepaspectratio=true]{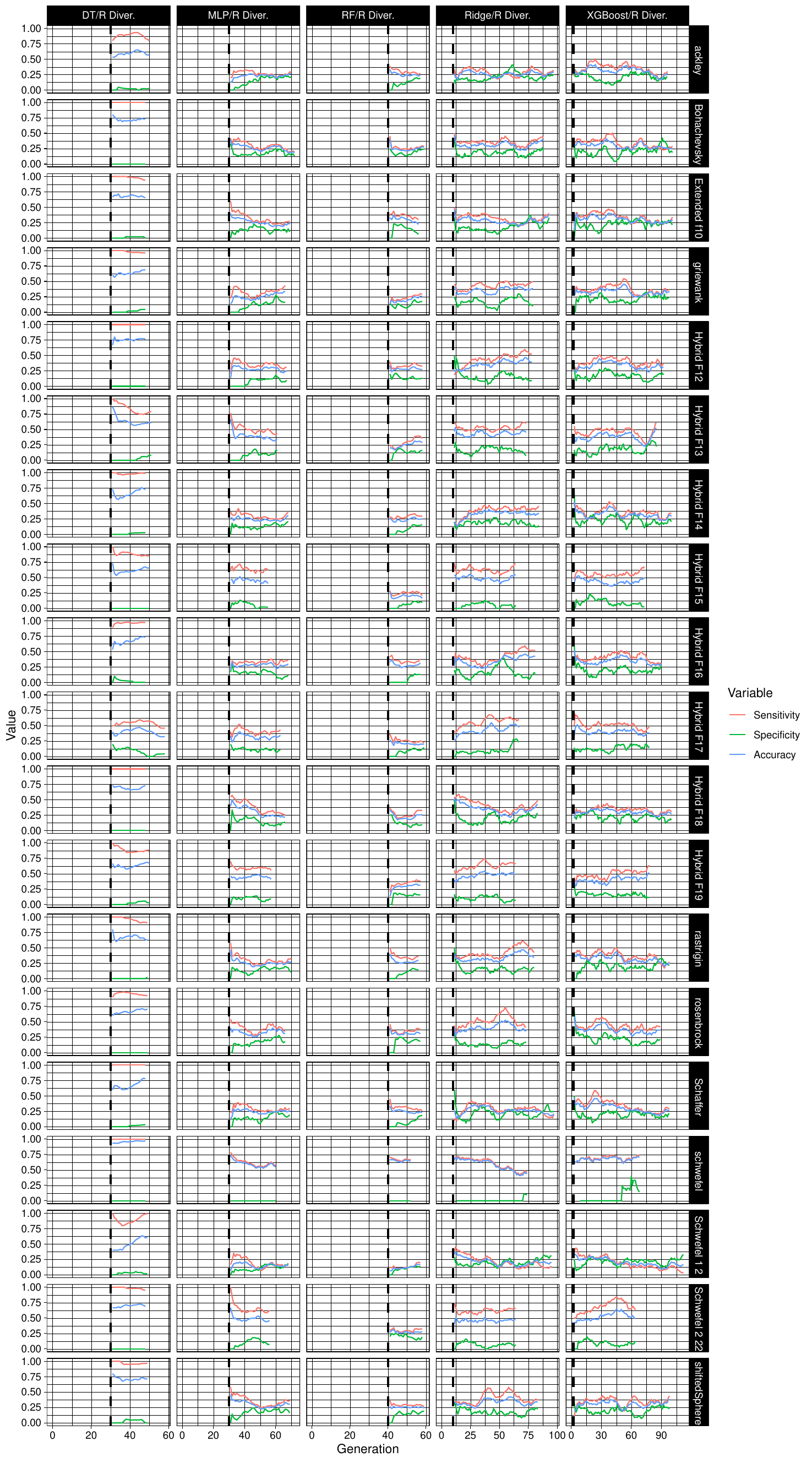}
  \caption{Surface sensitivity, specificity and accuracy. Vertical dashed lines define the point when warm up ends.}
  \label{app:fig:surfaceSens}
\end{figure*}

\begin{figure}[!b]
    \centering
  \includegraphics[width=\textwidth, height = 0.95\textheight, keepaspectratio=true]{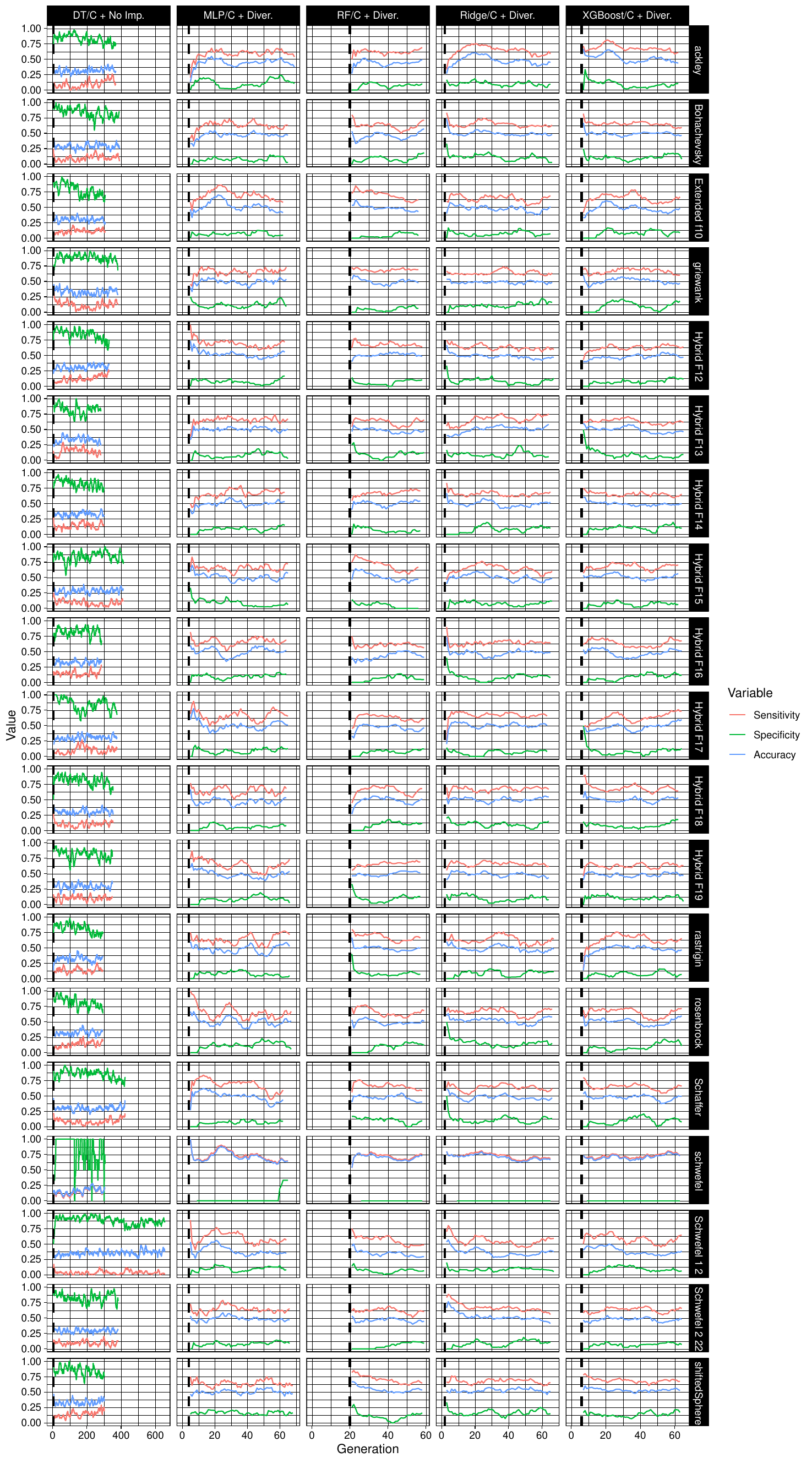}
  \caption{Pairwise sensitivity, specificity and accuracy. Vertical dashed lines define the point when warm up ends.}
  \label{app:fig:pairwiseSens}
\end{figure}

\begin{figure}[!b]
    \centering
  \includegraphics[width=\textwidth, height = 0.95\textheight, keepaspectratio=true]{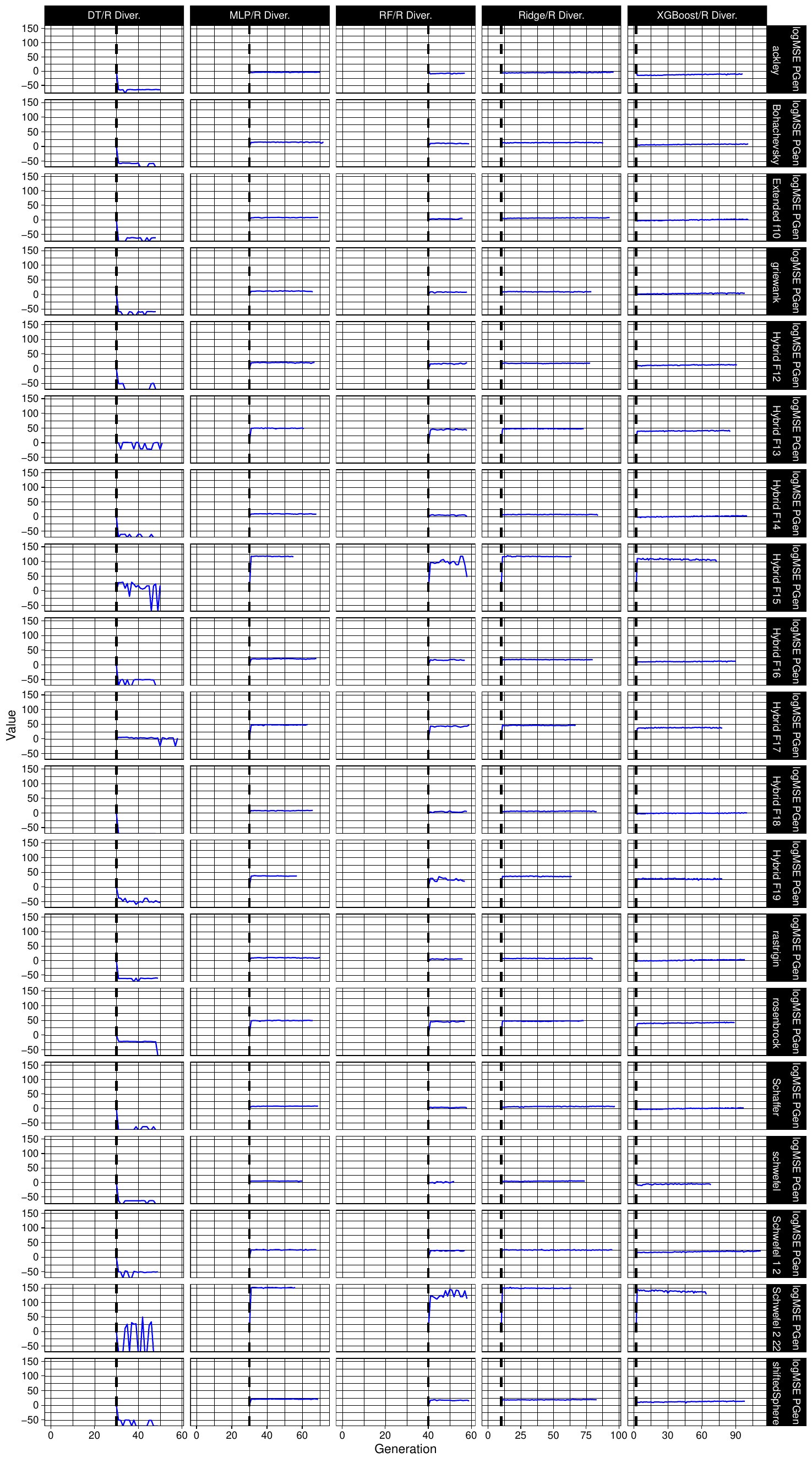}
  \caption{Surface Mean-Square Error analysis of previous generation. Vertical dashed lines define the point when warm up ends.}
  \label{app:fig:SurfaceMSEPrev}
\end{figure}

\begin{figure}[!b]
    \centering
  \includegraphics[width=\textwidth, height = 0.95\textheight, keepaspectratio=true]{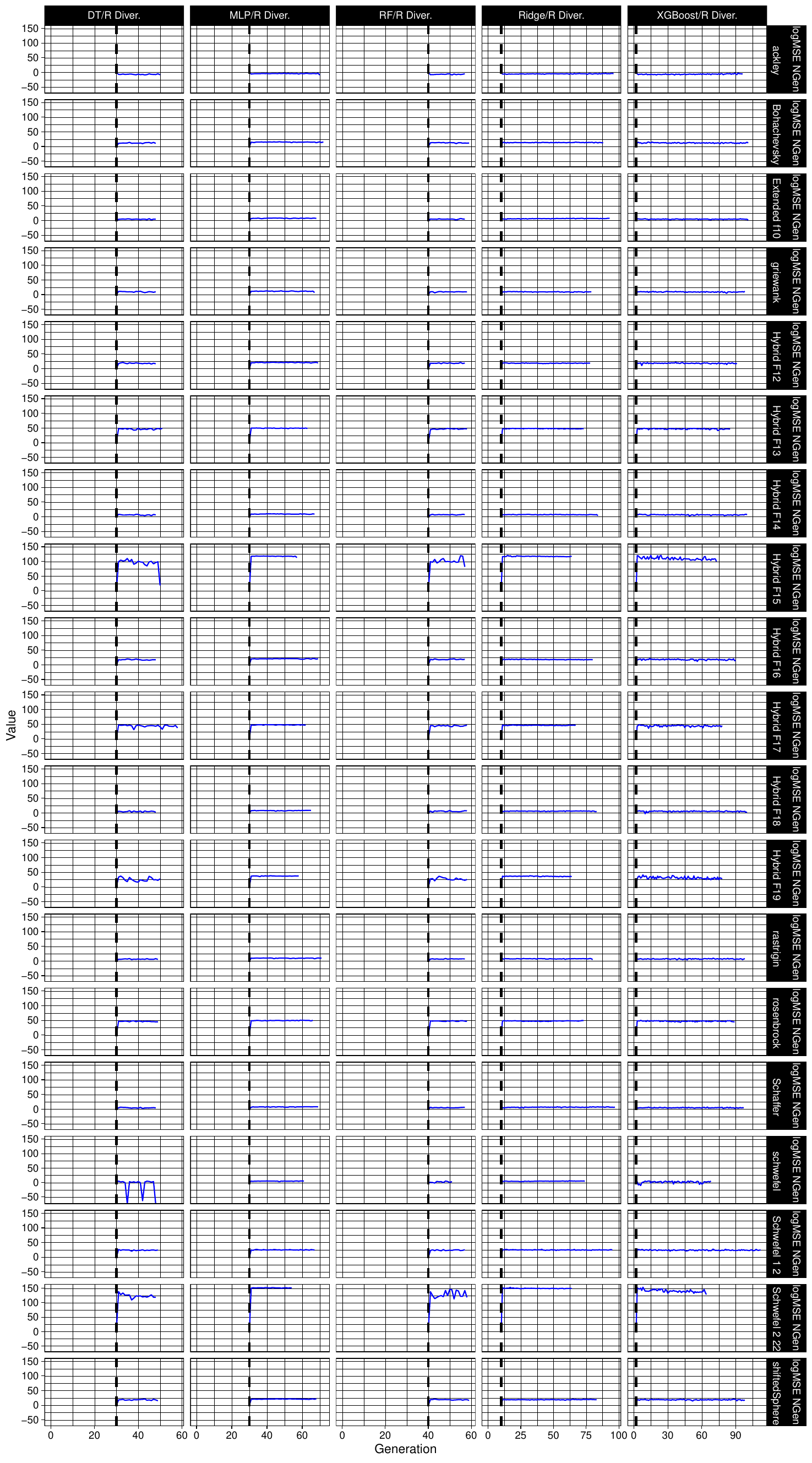}
  \caption{Surface Mean-Square Error analysis of next generation. Vertical lines dashed define the point when warm up ends.}
  \label{app:fig:SurfaceeMSENext}
\end{figure}

\FloatBarrier

\section{Pairwise Mapping Analysis}
In Section~\ref{sec:pairwise-mapping}, we introduce the \textbf{pairwise mapping function} that projects the pair of points from the original search space into a point in the pairwise training dataset. We proposed two different mapping alternatives, described in Equations~\ref{eq:mapping1} and \ref{eq:mapping2}.

\begin{figure*}[!htp]
    \centering
  \includegraphics[keepaspectratio=true, width=0.4\textwidth]{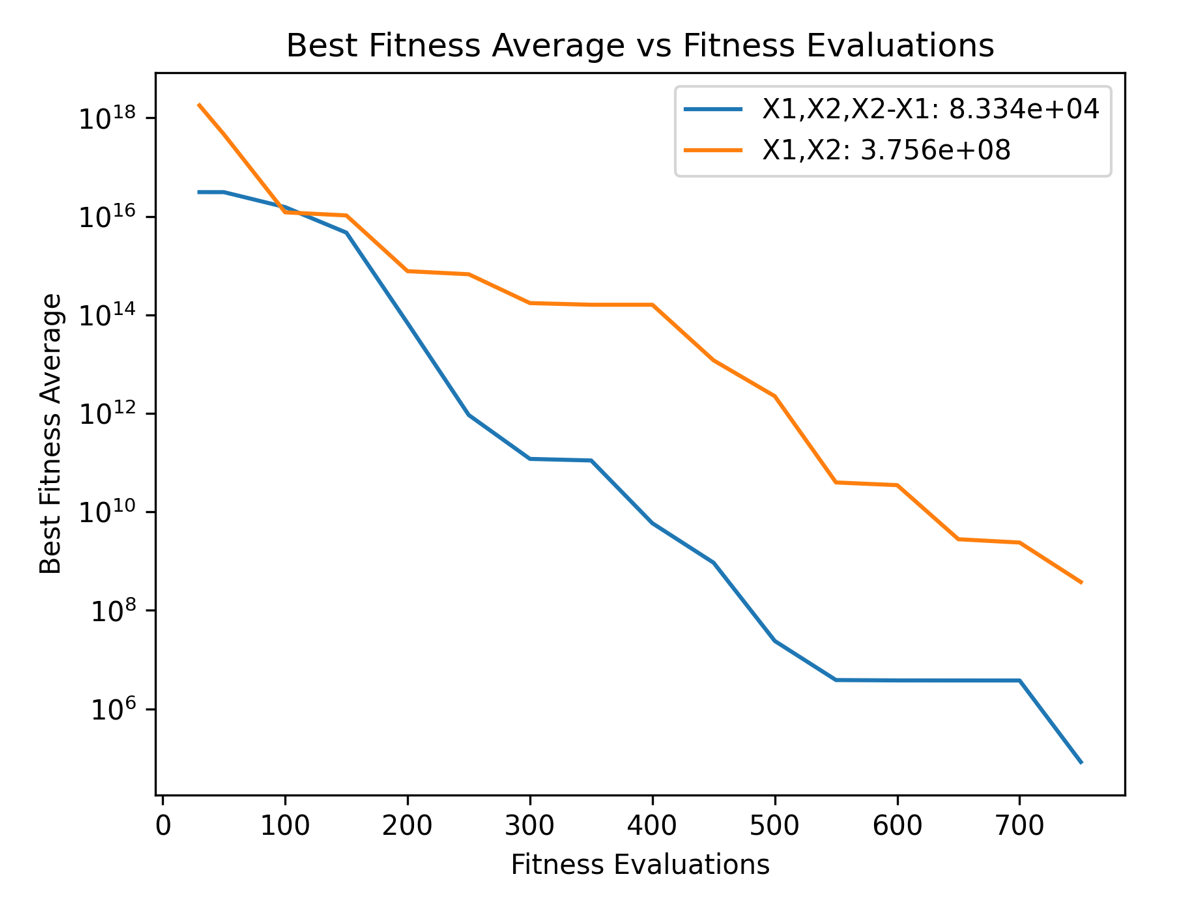}
  \caption{Comparison of average performance when using a DT/C pairwise trained with difference information (blue) and without this information (orange).}
  \label{rev:fig:differencetraining}
\end{figure*}

We have conducted a comparative analysis of the performance of these two mapping alternatives, and we have identified that, in a significant number of problems, the performance of the surrogate using Eq.~\ref{eq:mapping2} (that includes the difference between the two solutions) is better. Figure~\ref{rev:fig:differencetraining} shows an example of this different performance using Hybrid 15 function and a DT/C.

\end{document}